\documentclass[letterpaper,times]{IONconf}

\usepackage[dvipsnames]{xcolor}
\usepackage{url}

\usepackage[colorlinks=true,linkcolor=black,anchorcolor=black,citecolor=black,filecolor=black,menucolor=black,runcolor=black,urlcolor=black]{hyperref} 
\usepackage{amsmath}
\usepackage[]{graphicx}
\usepackage{float}
\usepackage{subcaption}
\usepackage{amssymb}
\usepackage[backend=biber,style=apa, natbib=true]{biblatex}
\usepackage{verbatim}
\usepackage{bm}
\usepackage{multirow}

\usepackage[ruled,vlined,linesnumbered]{algorithm2e}

\SetCommentSty{mycommfont}
\SetKwInput{KwInput}{Initialize} 
\SetKwInput{KwOutput}{Output at any $t$th time iteration} 

\usepackage{outlines}
\usepackage{enumitem}
\setenumerate[2]{label=\alph*.}
\setenumerate[3]{label=\roman*.}

\title{\huge Set-Valued Shadow Matching Using Zonotopes \\ for 3-D Map-Aided GNSS Localization}
\author{Sriramya Bhamidipati, Shreyas Kousik and Grace Gao,~\textit{Stanford University} }

\makeatletter 
\newcount\SOUL@minus
\makeatother  
\date{}

\begin{document}

\newcommand{\shrey}[1]{{\normalsize{{\color{WildStrawberry}(Shreyas:\ #1)}}}}
\newcommand{\ramya}[1]{{\normalsize{{\color{WildStrawberry}(Ramya:\ #1)}}}}

\newcommand{\R}{\mathbb{R}}
\newcommand{\N}{\mathbb{N}}

\newcommand{\regtext}[1]{\mathrm{\textnormal{#1}}}
\newcommand{\ts}[1]{\textsuperscript{#1}}
\newcommand{\defemph}[1]{\textbf{#1}}
\newcommand{\state}[1]{_{\regtext{#1}}}
\newcommand{\mc}[1]{\mathcal{#1}}

\newcommand{\zonofn}[1]{\regtext{zono}\!\left(#1\right)}
\newcommand{\ZSMfn}[1]{\regtext{ZSM}\!\left(#1\right)}
\newcommand{\LOSfn}[1]{\regtext{LOS}\!\left(#1\right)}
\newcommand{\polytopefn}[1]{\regtext{polytope}\!\left(#1\right)}
\newcommand{\vertfn}[1]{\regtext{vertices}\!\left(#1\right)}

\newcommand{\emptyarr}{[\ ]}
\newcommand{\zeros}{\mathbf{0}}
\newcommand{\ones}{\mathbf{1}}
\newcommand{\eye}{\mathbf{I}}

\newcommand{\inv}{^{-1}}
\newcommand{\comp}{^{\regtext{C}}}
\newcommand{\diag}[1]{\regtext{diag}\left(#1\right)}
\newcommand{\norm}[1]{\left\Vert#1\right\Vert}
\newcommand{\convhull}[1]{\regtext{convhull}\!\left(#1\right)}

\newcommand{\groundplane}{\mc{G}}
\newcommand{\AOI}{\mc{A}}
\newcommand{\sats}{\mc{S}}
\newcommand{\buildings}{\mc{B}}
\newcommand{\signalstrengths}{\mc{R}}
\newcommand{\signalstrength}{r}
\newcommand{\signalthreshold}{\bar{\signalstrength}\state{\tiny{LOS}}}
\newcommand{\shadow}[1]{S_{#1}}
\newcommand{\shadowVertex}[1]{C_{#1}}
\newcommand{\shadowvol}[1]{V_{#1}}
\newcommand{\shadowdir}[1]{\ell_{#1}}
\newcommand{\shadowdirzono}[1]{L_{#1}}
\newcommand{\shadowdirfn}[1]{\regtext{makeShadowDirection}\!\left(#1\right)}
\newcommand{\shadowdirzonofn}[1]{\regtext{makeShadowDirectionZono}\!\left(#1\right)}

\newcommand{\aoi}{_{\regtext{\tiny{AOI}}}}
\newcommand{\bldg}{_{\regtext{bldg}}}
\newcommand{\trng}{_{\regtext{trng}}}
\newcommand{\sat}{_{\regtext{sat}}}
\newcommand{\zono}{_{\regtext{zono}}}
\newcommand{\grnd}{_{\regtext{grnd}}}
\newcommand{\verts}{_{\regtext{vert}}}

\newcommand{\cno}{C/N_0}   
\maketitle
\pagestyle{plain}

\section*{biography}

\biography{Sriramya Bhamidipati}{is a postdoctoral scholar in the Aeronautics and Astronautics Department at Stanford University. She received her Ph.D. in Aerospace Engineering at the University of Illinois, Urbana-Champaign in 2021, where she also received her M.S in 2017. She obtained her B.Tech.~in Aerospace from the Indian Institute of Technology, Bombay in 2015. Her research interests include GPS, power and space systems, artificial intelligence, computer vision, and unmanned aerial vehicles.}

\biography{Shreyas Kousik}{is a postdoctoral scholar in the Department of Aeronautics and Astronautics at Stanford University.
He received his Ph.D. and M.S. in Mechanical Engineering at the University of Michigan - Ann Arbor, and his B.S. in Mechanical Engineering from the Georgia Institute of Technology.
His research focuses on verifiable perception, planning, and control for mobile robots.}

\biography{Grace Gao}{is an assistant professor in the Department of Aeronautics and Astronautics at Stanford University.
Before joining Stanford University, she was an assistant professor at University of Illinois at Urbana-Champaign. She obtained her Ph.D. degree at Stanford University. Her research is on robust and secure positioning, navigation and timing with applications to manned and unmanned aerial vehicles, robotics, and power systems.}
\section*{Abstract}
Unlike many urban localization methods that return point-valued estimates, a set-valued representation enables robustness by ensuring that a continuum of possible positions obeys safety constraints.
One strategy with the potential for set-valued estimation is GNSS-based shadow matching~(SM), where one uses a three-dimensional~(3-D) map to compute GNSS shadows~(where line-of-sight is blocked).
However, SM requires a point-valued grid for computational tractability, with accuracy limited by grid resolution.
We propose zonotope shadow matching (ZSM) for set-valued 3-D map-aided GNSS localization.
ZSM represents buildings and GNSS shadows using constrained zonotopes, a convex polytope representation that enables propagating set-valued estimates using fast vector concatenation operations.
Starting from a coarse set-valued position, ZSM refines the estimate depending on the receiver being inside or outside each shadow as judged by received carrier-to-noise density. 
We demonstrated our algorithm's performance using simulated experiments on a simple 3-D example map and on a dense 3-D map of San Francisco. 

\section{Introduction}\label{sec:intro}

\begin{figure}[H]
	\setlength{\belowcaptionskip}{-4pt}
	\centering
	\includegraphics[width=0.5\textwidth]{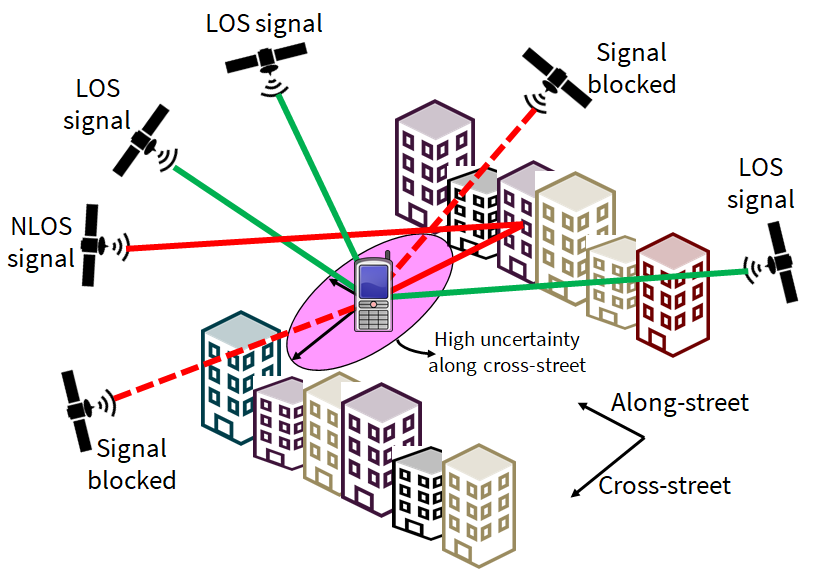}
	\caption{\small NLOS and multipath effects in GNSS satellites lead to high localization uncertainty in urban areas, especially along the cross-street direction. 
	The pink ellipse depicts uncertainty in the receiver position estimate.
	This figure is adapted from~\citep{Wang_2013}.}
	\label{fig:introduction}
\end{figure}

GNSS-based localization is often unreliable in dense urban areas.
As illustrated in Figure~\ref{fig:introduction}, direct line-of-sight~(LOS) GNSS signals can be blocked or reflected by tall buildings~\citep{Hofmann_Wellenhof_1992}, creating non-line-of-sight~(NLOS) and multipath effects, thereby lowering the number of visible GNSS satellites available for localization~\citep{Zhu_2018}.
In particular, the LOS satellite signals in the cross-street direction are more likely to be blocked or reflected by buildings than signals along the street.
Consequently, positioning accuracy is degraded in urban areas, especially in the cross-street direction.

Robustness to degraded accuracy is important for safety-critical GNSS applications such as autonomous driving and drone delivery.
As detailed next in our discussion of the related work, set-valued position estimates enable robustness guarantees by, e.g., ensuring an entire set of possible positions lies outside of obstacles or inside user-specified bounds.
Unfortunately, generating such position estimates is often challenging, requiring approximation with a discrete collection of points.
In this work, we propose a novel method of generating set-valued estimates without discretization by combining recent three-dimensional~(3-D) map-aided techniques from the GNSS community with recent geometric set representations from the controls community.

\subsection{Related Work}

A variety of approaches exist to address the positioning accuracy challenge by leveraging a 3-D map of an urban environment and buildings.
For example, in \defemph{ray tracing}, one can create discrete position candidates with a particle filter, and then find all the possible paths from each particle to the GNSS satellite, while considering a limit on the number of interactions with nearby buildings per each transmitted ray~\citep{Suzuki_2016,Miura_2013}. 
Essentially, ray tracing estimates the reflection route of NLOS signals and uses this to correct the bias in GNSS pseudoranges.
While there exist commercial implementations of this approach~\citep{iland_uber_rethinking_gps}, ray tracing is computationally-expensive and instead requires being offloaded to the cloud.
One strategy to reduce the computational complexity is to apply machine learning~(ML)-based GNSS to evaluate a pre-defined grid of position candidates with a neural network by collectively processing different heatmaps of pseudorange residuals~\citep{van_diggelen_android_improving_urban_gps}.
This enables real-time performance with server-based processing, but achieves onboard tractability by considering a tradeoff between the cross-street positioning accuracy and the computations executed to estimate the receiver position.

An alternative approach is \defemph{shadow matching~(SM)}~\citep{Groves_2011,Wang_2013,groves2015shadow_matching_challenges}, which can avoid offloading computation to the cloud~\citep{wang2013urban,Wang_2014}.
SM uses a 3-D map to estimate \defemph{GNSS shadows} of buildings, which are regions of the map where direct LOS signals are blocked.
A GNSS shadow is depicted in Figure~\ref{fig:GNSSshadow}.
The received carrier-to-noise density~($\cno$) of GNSS satellite signals enables determining if a receiver is inside or outside the GNSS shadow of each building.
However, to be computationally tractable, SM requires precomputing building skylines for a predefined grid of position candidates.
These skylines are compared against the azimuth and elevation of GNSS satellites at runtime \citep{groves2015shadow_matching_challenges,wang2013urban,Wang_2014}.
Each point-valued position candidate is evaluated by comparing the received $\cno$ with the precomputed building skyline to estimate a visibility score, which can be used to identify the most likely position candidate~(or candidates) and also a weighted empirical covariance for downstream use~\citep{wang2013urban}.

A common aspect of all the above methods is that they represent position estimates in a discretized way.
Since accuracy depends on the number of discrete points, these methods face a challenge in the case of \defemph{robust urban localization}, or ensuring that a position estimate obeys user-specified protection levels or safety bounds.
That is, one can use a finer discretization to achieve a more accurate estimate at the expense of more computation.

To avoid the point-valued discretization tradeoff, one can instead consider a \defemph{set-valued} approach to compute a \defemph{continuum} of state estimates given the initial set of states and known measurement bounds~\citep{Shi_2015,Shiryaev_2015,Combettes_1991,Scott_2016}.
This enables robustness by ensuring that the set lies within, e.g., user-specified safety bounds.
Recently, there has been growing interest in set-valued representations of receiver states and measurements for robust localization and motion planning in urban environments \citep{Shetty_2020,Bhamidipati_2020,Kousik_2019}.
Unfortunately, these methods typically require assuming a set-valued representation of uncertain measurements by, e.g., overapproximating a confidence level set of Gaussian distribution using a polytope \citep{Shetty_2020,Bhamidipati_2020}; in other words, one must make an assumption about the underlying distribution of measurements.
Therefore, it remains an open challenge to generate set-valued receiver position estimates for the distributions that are common in urban environments (e.g., multi-modal with disconnected components).

One strategy for modeling arbitrary set-valued state estimates is to use \defemph{zonotopes}.
Zonotopes are convex, symmetrical polytopes that can propagate set-valued state estimates using fast vector concatenation operations. 
Zonotopes are constructed as Minkowski sums~\citep{Althoff_2014} of line segments in an arbitrary-dimensional Euclidean space.
These objects are well-known for representing reachable sets of dynamical systems~\citep{althoff2016cora}, which can enable formally-verified motion planning, fault detection, and navigation~\citep{Shetty_2020,Bhamidipati_2020,Kousik_2019}.
These objects can also be extended to \defemph{constrained zonotopes}, which can represent arbitrary convex polytopes \citep{Scott_2016,raghuraman2020_cons_zono_ops}, avoiding the limitation of symmetry.
While zonotopes are widely used in robotics for path planning and collision avoidance, they have not been applied in the field of GNSS localization for addressing multipath/NLOS effects.

\subsection{Proposed Method}

\begin{figure}[ht]
	\setlength{\belowcaptionskip}{-4pt}
	\centering
	\includegraphics[width=0.6\textwidth]{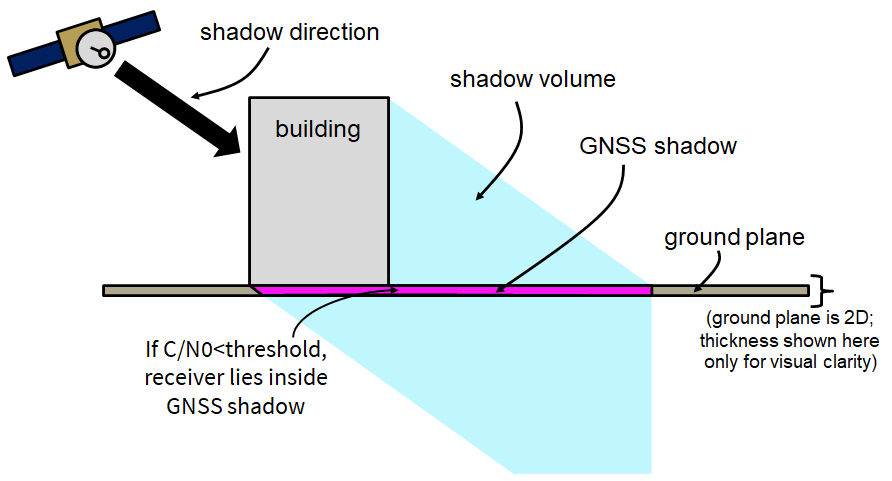}
	\caption{\small The intersection of a building's shadow volume (blue) with the ground plane (brown) produces a 2-D GNSS shadow (magenta); note, the 2-D ground plane is shown as having volume only for illustration's sake.
	The shadow volume is computed by extending the building along the shadow direction (black thick arrow) from the satellite to the building.
	Based on the received $\cno$, we evaluate the GNSS shadows to refine the set-valued receiver position estimate.
	In the proposed ZSM method, the building, shadow direction, shadow volume, and ground plane are all represented using constrained zonotopes, which enables efficient set-valued computation of GNSS shadows.}
	\label{fig:GNSSshadow}
\end{figure}

Our current work proposes a novel paradigm for 3-D map-aided GNSS localization based on GNSS shadow matching: set-based shadow matching termed as \defemph{Zonotope Shadow Matching~(ZSM)}.
This work is based on our recent ION GNSS+ 2021 conference paper~\citep{Bhamidipati_2021}.
We formulate buildings and GNSS shadows~(regions where LOS signals are blocked) using constrained zonotopes, wherein we compute the GNSS shadow for each satellite/building pair via the intersection of building shadow volume with that of ground plane. 
An illustration of these set operations is shown in Figure~\ref{fig:GNSSshadow}.
The ZSM algorithm begins with a coarse set-valued position estimate, which we iteratively refine using the GNSS shadow of each satellite/building pair in an urban map.
We use $\cno$ values to decide whether the receiver is inside or outside each GNSS shadow, and intersect the shadow with our position estimate if inside the shadow, or else subtract the shadow from our position estimate.
The final set-valued position estimate from our proposed ZSM algorithm can be incorporated into downstream polytopic estimation or verification methods without requiring assumptions about the distribution of the estimate.

A key feature of our proposed ZSM is that it avoids gridding and instead directly estimates a set-valued receiver position estimate by leveraging constrained zonotopes and two-dimensional~(2-D) polytopes (i.e., polygons).
Furthermore, given zonotope operations need fewer computations, our approach requires significantly lower compute power during offline step of shadow matching, while during online processing, the required compute power is on-par to that of conventional SM and lesser storage than that of conventional SM is incurred. 
Another advantage is that our formulation estimates a non-Gaussian distribution of possible receiver positions~(a continuum) in an urban environment, based on which many position statistics, such as worst-case error, multiple/distinct ambiguous modes, error bounds of each mode, can be evaluated in an \defemph{exact} manner. 

Given the recent availability of high-fidelity, open-source 3-D building maps~\citep{Miura_2015,Wang_2013}, we assume the inaccuracies in the building boundaries and road lanes, and the effect of building materials on GNSS signals to be negligible.
We also consider an ideal LOS/NLOS classifier, i.e., with true positive 100\%~(LOS detection) and true negative 100\%~(NLOS detection).
Our choice of an ideal LOS classifier is justified as it provides an uncorrupted platform to compare the performance of our proposed set-valued shadow matching (i.e., zonotope shadow matching - ZSM) with conventional SM.
Also note that, in this work, we do not explicitly account for multipath effects.
While this work demonstrates great success in achieving the exact bounds of set-valued receiver position for various simulation scenarios (see Section~\ref{sec:experiments}), ZSM suffers from some challenges similar to that of shadow matching, among which the key ones include brittleness to misclassification and inability to resolve multi-modal ambiguities without additional information. 
Note that addressing these other challenges, which are mentioned in a detailed manner in~\citep{groves2015shadow_matching_challenges}, is beyond the scope of this manuscript.
However, we have proposed extensions of this current work that account for brittleness to satellite misclassification in~\citep{neamati2022mosaic} and for resolving multi-modal ambiguities in~\citep{Neamati_2022}.
While our proposed ZSM proposed in this work cannot directly handle satellite misclassifications associated with the state-of-the-art, off-the-shelf classifiers~\citep{Xu2018classifier,xu2020machine}, we can still implement this work in real-world settings by leveraging a subset of GNSS satellites, either LOS or NLOS, that exhibit high satellite classifier probabilities (see experiments in Section~\ref{sec:experiments}).
Also, given that our proposed ZSM is an any-time implementation that works in a sequential manner, we can achieve real-time performance via strategic ordering of GNSS satellites, e.g., with an objective to maximize the Dilution of Precision~(DOP), so that the more useful satellites are processed first, and the algorithm is terminated when required to output the current estimate of the set-valued receiver position. 

\subsection{Contributions and Paper Organization}

Our key contributions are as follows: 
\begin{enumerate}
	\item We propose a novel set-valued approach to representing GNSS shadows using zonotopes.
	Offline, we represent buildings using constrained zonotopes.
	At runtime, we generate the GNSS shadow for each satellite/building pair by extending the building in the ``shadow direction'' from the satellite to the building, as shown in Figure~\ref{fig:GNSSshadow}; this operation is efficient due to our representation of buildings and GNSS shadows as constrained zonotopes.
	
	\item We propose the ZSM algorithm.
	Given an initial set-valued receiver position estimate, we iteratively subtract or intersect GNSS shadows across satellites to refine the estimate depending on the associated LOS/NLOS characteristics as judged by received $\cno$ values.
	To our knowledge, this is the first work utilizing set representations for 3-D map-aided GNSS.
	
    \item We experimentally demonstrate that ZSM computes a set-valued receiver position estimate, unlike conventional SM~\citep{Groves_2011,Wang_2013,groves2015shadow_matching_challenges,wang2013urban} for which point-valued position accuracy is limited by its precomputed grid resolution.
    We perform experiment simulations using i)~a simple 3-D map comprising two urban buildings and ii)~a dense 3-D building map of San Francisco. 
    We validate ZSM performance in terms of computation load (for both offline and online computations), point-valued estimation error and the position bounds~(size of the final set-valued receiver position estimate) in along-street and cross-street directions. 
\end{enumerate}

The remainder of the paper is organized as follows.
Section \ref{sec:preliminaries} introduces mathematical notations for the relevant set representations.
Section \ref{sec:proposed_zsm_algorithm} states our proposed ZSM algorithm.
Section \ref{sec:experiments} presents various experimental results that validate our proposed ZSM in simulation.
Finally, Section \ref{sec:conclusions_and_future_work} provides concluding remarks.
\section{Preliminaries of Set Representations}\label{sec:preliminaries}

We now introduce the mathematical notation and definitions used through this paper and present constrained zonotopes.

\subsection{Mathematical Notation and Definitions}\label{subsec:notation}

We denote the natural numbers as $\N$ and $n$-dimensional Euclidean space as $\R^n$.
Points and scalars are italic lowercase; sets and matrices/arrays are italic uppercase.
Sets of sets are script uppercase.
An $n\times n$ identity matrix is $\eye_n$.
An $n\times n$ array of zeros is $\zeros_n$, and a same-sized array of ones is $\ones_n$.
Similarly, an $n\times m$ array of zeros is $\zeros_{n\times m}$, and a same-sized array of ones is $\ones_{n\times m}$.
An empty vector or array is $\emptyarr$.

Let $A, B \subset \R^n$.
The \defemph{Minkowski (set) sum}~\citep{Althoff_2014} is $\oplus$, defined as $A\oplus B = \{a + b\ |\ a \in A, b \in B\}$.
For example, the Minkowski sum of two (non-parallel) line segments in $\R^2$ is a
convex polygon, whereas the Minkowski sum of two (non-parallel) polygons in $\R^3$ is a convex polyhedron.
The convex hull is $\convhull{A} = \{\lambda a_1 + (1-\lambda)a_2\ |\ a_1,a_2 \in A\}$.
A convex polytope $P \subset \R^n$ can be constructed as the convex hull of a set of vertices $V \subset \R^n$: $P = \convhull{V}$, which we call its \defemph{vertex representation.}

\subsection{Constrained Zonotopes}\label{subsec:constrained_zonotopes}

\begin{figure}[H]
	\setlength{\belowcaptionskip}{-4pt}
	\centering
	\begin{subfigure}[b]{0.49\textwidth}
	    \centering
		\includegraphics[width=0.93\textwidth]{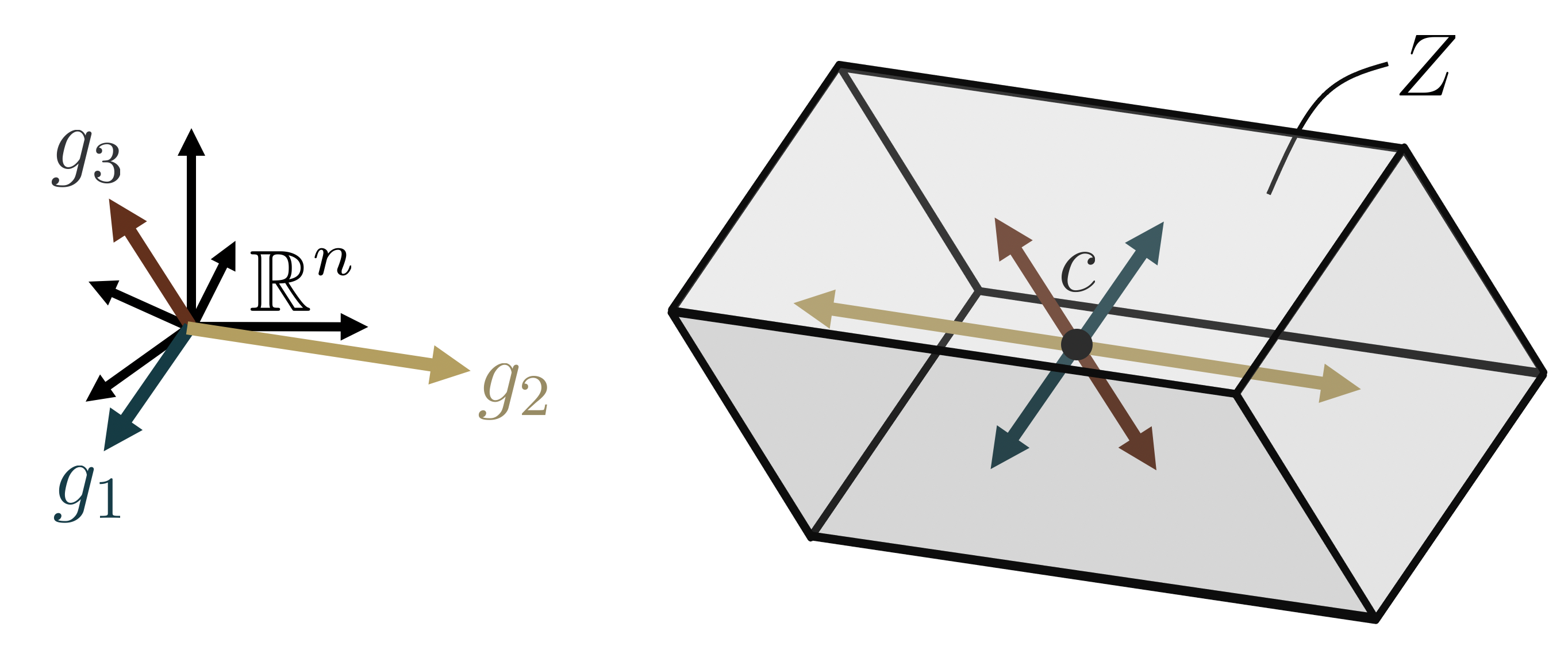}
	\caption{Example zonotope with three generators.}
	\label{fig:zono}
	\end{subfigure}
	\hspace{2mm}
	\begin{subfigure}[b]{0.49\textwidth}
	    \centering
	    \includegraphics[width=0.98\textwidth]{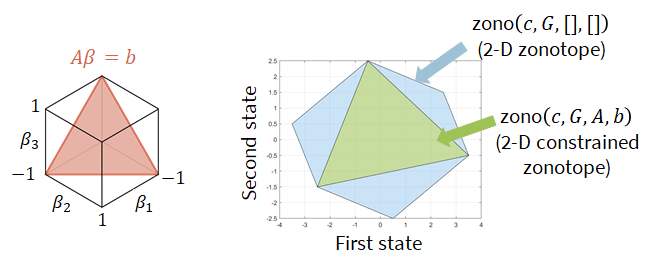}
	\caption{Example 2-D zonotope and 2-D constrained zonotope.}
	\label{fig:conzono}
	\end{subfigure}
	\caption{Subfigure (a) is a zonotope~(no constraints) with three generators, shown by grey volume: $Z = \zonofn{c,[g_1,g_2,g_3],\emptyarr,\emptyarr}$.
    The generators are shown both at the origin and extending from the zonotope's center to illustrate the concept that a zonotope is the Minkowski sum of the center with the line segments created by scaling each generator by all coefficient values in $[-1,1]$.
	Subfigure (b) shows an example illustration of a 2-D zonotope~(blue) and a 2-D constrained zonotope~(green).}
\end{figure}

We now define constrained zonotopes~\citep{Scott_2016}, which we use to represent buildings and GNSS shadows as shown in Figure~\ref{fig:GNSSshadow}.
Consider a \defemph{center} $c \in \R^n$, a \defemph{generator matrix} $G= [g_1,\cdots,g_m] \in \R^{n\times m}$, and linear constraints defined by $A \in \R^{p\times m}$ and $b \in \R^p$.
Per \citep{Scott_2016}, we define a \defemph{constrained zonotope} $Z \subset \R^n$ as a set
\begin{align}\label{eq:constrained_zonotope}
    Z = \zonofn{c,G,A,b} = \left\{c + G\beta \in \R^n |\ \beta \in [-1,1]^m\ \regtext{and}\ A\beta = b \right\} \subset \R^n.
\end{align}
The columns of $G$ are called \defemph{generators}.
Also note that a \defemph{zonotope} is a constrained zonotope with no constraints defined, which we denote by $\zonofn{c,G,\emptyarr,\emptyarr} = \{c + G\beta \in \R^n\ |\ \beta \in [-1,1]^m\}.$
An example 3-D zonotope~(i.e., no constraints) is shown in Figure~\ref{fig:zono}.

For another example illustration with $c=\begin{bmatrix} 0 \\ 0\end{bmatrix}$, $G=\begin{bmatrix}
1.5 & -1.5 & 0.5 \\
1 & 0.5 & -1
\end{bmatrix}$, $A=[1,1,1]$ and $b=1$, Figure~\ref{fig:conzono} shows the 2-D zonotope~$\zonofn{c,G,\emptyarr,\emptyarr}$ in blue while the 2-D constrained zonotope~$\zonofn{c,G,A,b}$ is shown in green.
\section{Proposed ZSM Algorithm} \label{sec:proposed_zsm_algorithm}

Our proposed ZSM algorithm is summarized as follows.
First, we create a 3-D ``shadow volume'' for each satellite/building pair by extending the building in the ``shadow direction'' from the satellite to the building.
Note, the building, shadow direction, and shadow volume are all represented as constrained zonotopes.
Second, we intersect the shadow volumes with the ground plane (on which we assume the receiver is positioned) to compute GNSS shadows as 2-D constrained zonotopes.
Third, by comparing received $\cno$ values against a user-specified threshold, we either subtract (if LOS signal) or intersect (if NLOS signal) each GNSS shadow with our current set-valued receiver position estimate.
Starting from an initial set of position estimates, the final position estimate set is obtained by iterating through all buildings and satellites.

We note that this is a \defemph{snapshot} algorithm, meaning that we make use of only the instantaneous data available at any given time instance.
Furthermore, the order in which GNSS satellites and buildings are considered does not affect the final set-valued estimate, meaning that the method is parallelizable.
However, we leave extensions to time-varying shadows, receiver motion models, and parallelization as future work.
To proceed, we describe our model and assumptions of the urban environment and GNSS satellite signals, then detail how we compute shadows, and finally present the ZSM algorithm.

\subsection{System Model}

\subsubsection{Urban Environment Map}\label{subsec:urban_map_assumptions}

We assume access to a 3-D building map of the urban environment in which we plan to perform positioning, as is done in the related work \citep{Adjrad_2017,van_diggelen_android_improving_urban_gps,iland_uber_rethinking_gps}.
We assume such a map consists of a ground plane and a collection of buildings.

We assume that the \defemph{ground plane}, denoted $\groundplane$, is either a single 2-D constrained zonotope (as in Figure~\ref{fig:GNSSshadow}), or a collection of 2-D constrained zonotopes: $\groundplane = \{G_k\}_{k=1}^{n\grnd}$, where each $G_k$ is a 2-D constrained zonotope and $n\grnd \in \N$.
Thus, our algorithm does not require the ground plane to be constant within an area of interest. 
We can account for the variations in the ground height by modeling the ground plane as a collection of triangles, wherein each triangle is converted to a constrained zonotope, and the ground plane is represented by a collection of constrained zonotopes.
There exist many standard ways to extract ground height, of which one straightforward way is to extract the ground plane information from the 3-D building map being utilized; and the other is to leverage the open-source information on digital terrain models~(DTM), digital surface models~(DSM) or digital elevation models~(DEM)~\citep{Chen_2017}.
We also assume the ground plane is bounded (which is implied by $n\grnd$ being finite).
In particular, we only consider a specific, bounded urban area that could be potentially identified by a coarse position estimate to within a few kilometers of the localization error.

We denote the set of all \defemph{buildings} as $\buildings = \{B_i\}_{i=1}^{n\bldg}$, where $n\bldg \in \N$ is finite and $B_i \subset \R^3$ denotes the $i$\ts{th} building.
We further assume that each building can be represented as a union of constrained zonotopes:
\begin{align}\label{eq:assum_buildings_are_conzonos}
    B_i = \bigcup_{l=1}^{n_i} Z_l
    \quad\regtext{with each}\quad
    Z_l = \zonofn{c_{i,l},G_{i,l},A_{i,l},b_{i,l}},
\end{align}
where $n_i$ is the number of constrained zonotopes representing the building~$B_i$.
Note that any convex polytope is a constrained zonotope \citep{Scott_2016}, so this is not a restrictive assumption.
Also note that, since the union of constrained zonotopes cannot be represented as a constrained zonotope (because it is not necessarily a convex polytope), in practice we represent each building as a list of constrained zonotopes (i.e., the union is treated implicitly).

\subsubsection{Preprocessing Standard 3-D Maps}\label{subsubsec:3-D_map_vertex_rep}

Representing buildings as constrained zonotopes requires preprocessing a 3-D map.
Standard 3-D maps generated by computer-aided design software are often represented by a \defemph{triangulation}, or a union of triangles.
That is, for a building $B_i \in \buildings$, 
\begin{align}\label{eq:building_are_triangles}
    B_i = \bigcup_{l=1}^{n\trng} T_l\quad
    \regtext{with each}\quad
    T_l = \convhull{\{t_{l,1},t_{l,2},t_{l,3}\}}
\end{align}
such that $n\trng \in \N$ is finite and each $t_{l,j} \in \R^3$ is a vertex of the triangle.
Assuming our map is in such a standard format, we preprocess a 3-D map as follows to satisfy our assumption in \eqref{eq:assum_buildings_are_conzonos}.

First, note that the convex hull of constrained zonotopes is given by~\cite[Theorem 5]{raghuraman2020_cons_zono_ops} as follows.
Let $Z_1 = \zonofn{c_1,G_1,A_1,b_1}\subset \R^n$ and $Z_2 = \zonofn{c_2,G_2,A_2,b_2} \subset \R^n$.
Suppose that $G_1 \in \R^{n\times m_1}$, $G_2 \in \R^{n\times m_2}$, $A_1 \in \R^{k_1\times m_1}$, and $A_2 \in \R^{k_2\times m_2}$.
Then
\begin{subequations}\label{eq:conzono_convhull}
\begin{align}
    \convhull{Z_1 \cup Z_2} &= \zonofn{\tfrac{1}{2}(c_1 + c_2),\ [G_1, G_2, \tfrac{1}{2}(c_1 - c_2), \zeros],\ A\state{CH},\ b\state{CH}},\quad\regtext{where}\\
    A\state{CH} &= \begin{bmatrix}
        A_1 & \zeros_{k_1\times m_2} & -\tfrac{1}{2}b_1 & \zeros_{k_1\times 2(m_1+m_2)} \\
        \zeros_{k_2\times m_1} & A_2 &  \tfrac{1}{2}b_2 & \zeros_{k_2\times 2(m_1+m_2)} \\
        A_{3,1} & A_{3,2} & A_{3,0} & \eye_{2(m_1+m_2)}
    \end{bmatrix},\quad
    b\state{CH} = \begin{bmatrix}
        \tfrac{1}{2}b_1 \\
        \tfrac{1}{2}b_2 \\
        -\tfrac{1}{2}\ones_{2(m_1+m_2)\times 1}
    \end{bmatrix},\\
    A_{3,1} &= \begin{bmatrix}
        \eye_{m_1} \\ -\eye_{m_1}\\ \zeros_{2m_2\times m_1}
    \end{bmatrix},\quad
    A_{3,2} = \begin{bmatrix}
        \zeros_{2m_1\times m_2} \\ \eye_{m_2} \\ -\eye_{m_2}
    \end{bmatrix},\quad\regtext{and}\quad
    A_{3,0} = \begin{bmatrix}
        -\tfrac{1}{2}\ones_{2m_1\times 1} \\ +\tfrac{1}{2}\ones_{2m_2\times 1}
    \end{bmatrix}.\quad
\end{align}
\end{subequations}

Now, notice that each vertex can be represented as a constrained zonotope, $t_{l,j} = \zonofn{t_{l,j},\emptyarr,\emptyarr,\emptyarr}$.
So, we can represent $T_l$ as a constrained zonotope by taking $\convhull{\convhull{\{t_{l,1},t_{l,2}\}},\{t_{l,3}\}}$, meaning that we apply \eqref{eq:conzono_convhull} twice.
Thus, we preprocess a 3-D map by iterating over all of its triangles and representing each one as a constrained zonotope.
In this case, $n\trng=n_i$. 
Note that another series of convex hull operations can be performed to combine the constrained zonotopes~(each corresponds to a triangle) to form a single constrained zonotope associated with each building, in which case $n_i=1$.

\subsubsection{Receiver Assumptions}\label{subsec:receiver_assumptions}

We assume that the receiver is on the 2-D ground plane, which we find reduces the computational burden, and is common in conventional SM~\citep{groves2015shadow_matching_challenges}.
We plan to extend this method to 3-D as future work.

To create a set-valued receiver position estimate, we consider an \defemph{area of interest} (AOI):
\begin{align}
    \AOI = \{A_k\}_{k=1}^{n\aoi} \subseteq \groundplane,
\end{align}
where $A_k$ is a single ground plane constrained zonotope.
Since the ground plane is bounded, the AOI is bounded.
While it is reasonable to treat the entire ground plane as the AOI, we make this distinction to enable our algorithm to take advantage of the non-shadow-based localization data obtained from various information sources. 
Note that the requirements for determining size of the AOI remain the same as that for conventional SM, i.e., the AOI needs to cover the user's true location. 
Some potential methods for this include the following, among which a few are explained in a more detailed manner in~\citep{groves2015shadow_matching_challenges}: a)~Use of position estimate and uncertainty bounds from the conventional GNSS ranging solution, which is based on weighted analysis of all GPS satellites or only the LOS ranging signals, b)~Use of other sensors, such as camera-based geotagging, WiFi-based positioning, sensor fusion of GPS-vision or GPS-LiDAR, multi-agent system with inter-ranging, c)~Use of predicted position estimate and uncertainty from onboard state estimation techniques, such as Kalman filter, particle filter, and d)~Use of a heuristic approach that involves either considering a very large AOI and later narrowing down the correct mode from among the multiple ambiguous modes of shadow matching identified or starting with an initial AOI and increasing the size by a scale factor until a viable set-estimate of receiver position is obtained. 
Furthermore, the user can choose to eliminate the segments in the AOI that lie within building footprints since the user is expected to be outdoors and not inside the building. 

\subsubsection{Satellites and Received $\cno$}\label{subsec:satellite_assumptions}

We denote the observed set of GNSS satellites by $\sats = \{s_j\}_{j=1}^{n\sat}$, where each $s_j \in \R^3$ is a satellite location and $n\sat$ is the total number.
We obtain the received $\cno$ values from the GNSS receiver; we denote these values by $\signalstrengths = \{\signalstrength_j\}_{j=1}^{n\sat}$, one for each satellite.
In this work, we assume an ideal LOS/NLOS classifier, and thus do not need to account for the associated false alarm or missed-detection rates.

\subsection{Computing Shadows}\label{subsec:computing_shadows}

Now we detail how we compute shadows using operations on constrained zonotopes.

\subsubsection{Overview and Assumptions}

To create shadows, first, we consider the \defemph{shadow direction} associated with each satellite/building pair.
This is a unit vector pointing in the direction from the satellite's position to a point in the building.
Then, we compute the 3-D \defemph{shadow volume} for the satellite/building pair by extending the building in the shadow direction.
Finally, we compute the 2-D \defemph{GNSS shadow} for the satellite/building pair, which is a subset of the AOI in which the receiver does or does not lie depending on the received $\cno$.

Note, we have made a key assumption that the shadow direction can be treated as a single direction, as opposed to a cone of possible directions, because the distance from the satellite to any building is much larger than the size of the building.
However, for a more accurate model, or if the satellite is at a low elevation angle that would produce a large shadow, we can instead consider a cone of shadow directions created by the convex hull of $s_j$ and $B_i$ using \eqref{eq:conzono_convhull}.
Since this convex hull strategy requires a more complicated presentation of the algorithm, one can discard low-elevation satellites while implementing ZSM.

\subsubsection{Computing Shadow Directions}

We compute the \defemph{shadow direction} $\shadowdir{i,j}$ as a unit vector from the satellite position $s_j$ to a point $b_i \in B_i$.
Recall from \eqref{eq:assum_buildings_are_conzonos} that each building is represented as a union of constrained zonotopes.
Recall also that we assume the shadow direction is the same for any $b_i \in B_i$.
So, we choose $b_i$ as the mean of the vertices of all of the constrained zonotopes comprising $B_i$.

To get vertices of $B_i$, we use the following procedure.
Recall that $B_i$ is a union of constrained zonotopes.
Using techniques from \citep{althoff2016cora,vert2lcon}, one can convert a constrained zonotope $Z = \zonofn{c,G,A,b}\subset \R^n$ to a vertex representation as follows.
Suppose $Z$ has $m$ generators (i.e., $G \in \R^{n\times m}$).
Notice from \eqref{eq:constrained_zonotope} that $Z$ is an affine map of the polytope $P = \{x \in \R^m\ |\ \norm{x}_{\infty} \leq 1\ \regtext{and}\ Ax = b\}$.
That is we can write, $Z = c + GP$.
To get a vertex representation of $Z$, we first apply \citep{vert2lcon} to enumerate the vertices of $P$ get a set $W \subset \R^m$ for which $P = \convhull{W}$.
Then, the vertices of $Z$ are vertices of the set $\convhull{\{c + Gw\ |\ w \in W\}} \subset \R^n$, which we compute using \citep{althoff2016cora}.

We summarize the above procedure with a function that applies to the \textit{union} of constrained zonotopes comprising $B_i$.
Let $\mc{Z} = \{Z_i\}_{i=1}^{n_Z}$ be a set of constrained zonotopes.
Then
\begin{align}\label{eq:vert_fn}
    \{v_i\}_{i=1}^{n\state{verts}} = \vertfn{\mc{Z}}
\end{align}
is a vertex representation of the polytope (not necessarily convex) created by the union $\bigcup_{i=1}^{n_Z} Z_i$.

Now, consider the set $\{v_k\}_{k=1}^{n\verts} = \bigcup_{l=1}^{n_i} \vertfn{Z_l}$, where $Z_l$ is as in \eqref{eq:assum_buildings_are_conzonos}.
We estimate the point $b_i$ as $b_i = \frac{1}{n\verts}\sum_{k=1}^{n\verts} v_k$.
Finally, we compute the shadow direction as
\begin{align}\label{eq:shadow_dir}
    \shadowdir{i,j} = \frac{s_j - b_i}{\norm{s_j - b_i}_2}.
\end{align}
Next, we use $\shadowdir{i,j}$ to compute a building's shadow as a volume in 3-D space.

\subsubsection{Computing Shadow Volumes}

To compute a \defemph{shadow volume} for satellite/building pair $(i,j)$, we extend $B_i$ in the shadow direction $\shadowdir{i,j}$.
This extension is achieved by taking the Minkowski sum of the building and the shadow direction.

Since the building and shadow direction are constrained zonotopes, to proceed, we now define the Minkowski sum of constrained zonotopes.
Consider the example zonotopes $Z_1 = \zonofn{c_1,G_1,A_1,b_1}$ and $Z_2 = \zonofn{c_2,G_2,A_2,b_2} \subset \R^n$.
Suppose that $G_1 \in \R^{n\times m_1}$, $G_2 \in \R^{n\times m_2}$, $A_1 \in \R^{k_1\times m_1}$, and $A_2 \in \R^{k_2\times m_2}$.
Note, $k_1$ is the number of constraints for $Z_1$, and similarly $k_2$ for $Z_2$, as per \eqref{eq:constrained_zonotope}.
Per \citep{Scott_2016}, the Minkowski sum of constrained zonotopes is
\begin{align}\label{eq:zono_mink_sum}
Z_1 \oplus Z_2 = \zonofn{c_1 + c_2, [G_1, G_2], A_\oplus, b_\oplus},\quad
A_\oplus = \begin{bmatrix}
    A_1 & \zeros_{k_1\times m_2} \\
    \zeros_{k_2\times m_1} & A_2
\end{bmatrix},\quad
b_\oplus = \begin{bmatrix}
    b_1 \\ b_2
\end{bmatrix},
\end{align}
Note, this follows from the definition of constrained zonotopes.

Now we can compute the shadow volume.
First, consider the zonotope
\begin{align}\label{eq:shadow_dir_zonotope}
    \shadowdirzono{i,j} = \zonofn{\zeros_{3\times 1},\ \epsilon\cdot\shadowdir{i,j}, \emptyarr, \emptyarr},
\end{align}
where $\epsilon$ denotes a scaling factor that is greater than at least the height of the tallest building in the set~$\buildings$ (in practice we use $\epsilon = 10^5$ meters).
In other words, $\shadowdirzono{i,j}$ is a zonotope representing a long line segment extending in the shadow direction $\shadowdir{i,j}$.
Then, the shadow volume of building $B_i$ with respect to satellite $j$ is
\begin{align}\label{eq:shadow_volume}
    \shadowvol{i,j} = B_i \oplus \shadowdirzono{i,j}
    = \bigcup_{l=1}^{n_i} \{Z_l \oplus \shadowdirzono{i,j}\},
\end{align}
where $Z_l$ are the constrained zonotopes comprising $B_i$ as in \eqref{eq:assum_buildings_are_conzonos}.
In practice, we represent the union of constrained zonotopes as a list of constrained zonotopes; so, we hold on to each
\begin{align}\label{eq:shadow_volume_l}
    \shadowvol{l} = Z_l \oplus \shadowdir{i,j},    
\end{align}
where we have omitted the $i,j$ indices from $\shadowvol{l}$ for ease of notation.

\subsubsection{Computing GNSS Shadows}

Finally, to compute a 2-D \defemph{GNSS shadow} for the satellite/building pair $(i,j)$, we intersect the shadow volume with the 2-D AOI.

Recall that the shadow volume and the AOI are represented by constrained zonotopes.
Again let $Z_1 = \zonofn{c_1,G_1,A_1,b_1}$ and $Z_2 = \zonofn{c_2,G_2,A_2,b_2} \subset \R^n$ with $G_1 \in \R^{n\times m_1}$, $G_2 \in \R^{n\times m_2}$, $A_1 \in \R^{k_1\times m_1}$, and $A_2 \in \R^{k_2\times m_2}$.
Per \cite[Prop. 1]{Scott_2016}, the intersection of constrained zonotopes is a constrained zonotope:
\begin{align}
Z_1 \cap Z_2 = \zonofn{c_1,[G_1, \zeros_{n\times m_2}],A_\cap, b_\cap},\quad
A_\cap = \begin{bmatrix}
    A_1 & \zeros_{k_1\times m_2} \\
    \zeros_{k_2\times m_1} & A_2 \\
    G_1 & - G_2
\end{bmatrix}\quad\regtext{and}\quad
b_\cap = \begin{bmatrix}
    b_1 \\ b_2 \\ c_2 - c_1
\end{bmatrix}.\label{eq:zono_int}
\end{align}

Therefore, for the shadow volume $\shadowvol{l}$ and area of interest $\AOI = \{A_k\}_{k=1}^{n\aoi}$, we can compute a \defemph{shadow} $\shadow{i,j}$ as
\begin{subequations}\label{eq:GNSS_shadow}
\begin{align}
    \shadow{i,j} &= \bigcup_{k=1}^{n\aoi} \left(\shadowvol{l}\cap A_k \right)\\
        &= \bigcup_{k=1}^{n\aoi}\bigcup_{l=1}^{n_i} \left(\{Z_l \oplus \shadowdirzono{i,j}\} \cap A_k \right),
\end{align}
\end{subequations}
where we used \eqref{eq:shadow_volume} to expand $\shadowvol{i,j}$ into a union of constrained zonotopes.
As mentioned earlier, we represent a union of constrained zonotopes as a list of constrained zonotopes.
So, in practice, we consider each $k$\ts{th} shadow,
\begin{align}\label{eq:shadow_k}
    \shadow{k} = \left(Z_l \oplus \shadowdirzono{i,j}\right) \cap A_k,
\end{align}
which is a single constrained zonotope corresponding to each $A_k$; we omit the $i,j,l$ indices for ease of notation.

\subsection{Algorithm Details} \label{sec:algorithm_details}

We now describe how to initialize and perform ZSM, summarized in Algorithm~\ref{alg:zono_shadow_matching}.
This is a modification of the conventional SM approach originally introduced in \citep{Groves_2011}; in particular, we enable set-valued receiver position estimates.
We emphasize again that this is a snapshot method, considering only the data available at a single time instance.

\IncMargin{1em}
\begin{algorithm}[ht]
\caption{$P = \ZSMfn{\buildings,\sats,\signalstrengths,\signalthreshold,\AOI,\epsilon}$ // zonotope shadow matching (Snapshot)}\label{alg:zono_shadow_matching}

{/* Inputs:
\begin{itemize}
    \item Urban 3-D map of buildings $\buildings = \{B_i\}_{i=1}^{n\bldg}$ indexed by $i$, with each $B_i$ as a set of constrained zonotopes $\{Z_l\}_{l=1}^{n_i}$
    \item Observed SVs $\sats = \{s_j\}_{j=1}^{n\sat}$ indexed by $j$
    \item Received $\cno$ values $\signalstrengths = \{\signalstrength_j\}_{j=1}^{n\sat}$ for each satellite
    \item Prespecified $\cno$ threshold $\signalthreshold$
    \item Area-of-interest $\AOI = \{A_k\}_{k=1}^{n\aoi}$ indexed by $k$ and represented as a set of constrained zonotopes
    \item Shadow length scale $\epsilon \approx 10^5$ m.
\end{itemize}
*/}

$P \leftarrow \vertfn{\AOI}$ // initialize coarse set-valued position estimate as a polygon~(2-D polytope) \label{line:init_P}

\For{each $s_j \in \sats$\label{line:for_sat}}{
    $\shadowVertex{k} = \emptyset$ // initialize 2-D GNSS shadow as an empty polytope in vertex representation \\
    \For{each $B_i \in \buildings$\label{line:for_bldg}}{
        $\shadowdir{i,j} \leftarrow \shadowdirfn{B_i,s_j}$ // make shadow direction as in \eqref{eq:shadow_dir}\label{line:shadow_dir}\\
    
        $\shadowdirzono{i,j} \leftarrow \zonofn{\zeros_{3\times 1},\ \epsilon\cdot\shadowdir{i,j}, \emptyarr, \emptyarr}$ // make shadow direction zonotope as in \eqref{eq:shadow_dir_zonotope}\\
        
        \For{each $Z_l \in B_i$}{
            $\shadowvol{l} \leftarrow Z_l \oplus \shadowdirzono{i,j}$ // construct shadow volume zonotope as in \eqref{eq:shadow_volume_l}\label{line:make_shdw_vol_zono} \\
            
            \For{each $A_k \in \AOI$}{
                $\shadow{k} \leftarrow \shadowvol{l} \cap A_k$ // create 2-D GNSS shadow using zonotope intersection as in \eqref{eq:shadow_k}\label{line:make_gnss_shdw} \\
                
                $\shadow{k} \leftarrow \vertfn{\{\shadow{k}\}}$ // convert 2-D GNSS shadow to vertex representation with \eqref{eq:vert_fn} \label{line:conv_gnss_shdw_to_vtx} \\
                $\shadowVertex{k} \leftarrow \shadowVertex{k} \cup \shadow{k}$ // concatenate vertex-represented regions of 2-D GNSS shadow across buildings \label{line:concatenate_vtx_gnss_shdw}\\
            }
        }
    }            
    \eIf{$\signalstrength_j < \signalthreshold$\label{line:if_NLOS}}{
        $P \leftarrow P \cap \shadowVertex{k}$ // in shadow (NLOS), so intersect shadow with estimated position\label{line:shdw_intersect}
    }{\label{line:else_LOS}
        $P \leftarrow P \setminus \shadowVertex{k}$ // not in shadow (LOS), so subtract shadow from estimated position\label{line:shdw_subtract}
    }
}
\Return $P$ // updated set-valued receiver position estimate \label{line:return}
\end{algorithm}
\DecMargin{1em}

\subsubsection{Algorithm Overview}

We begin with the entire AOI $\AOI \subseteq \groundplane$ as an initial 2-D set-valued estimate of our receiver position (Line \ref{line:init_P}).
If we have no localization information available besides the fact that we are in an urban area, then we set $\AOI = \groundplane$.
Then, for each GNSS satellite (Line \ref{line:for_sat}), we perform the following steps:
\begin{enumerate}
    \item First, for each building (Line \ref{line:for_bldg}), we compute the 3-D shadow volume as in \eqref{eq:shadow_volume} (Lines \ref{line:shadow_dir}--\ref{line:make_shdw_vol_zono}). Then, we intersect the shadow volume with the AOI to find a 2-D GNSS shadow (Line \ref{line:make_gnss_shdw}), which we efficiently convert to a generic vertex representation (Line \ref{line:conv_gnss_shdw_to_vtx}).
    
    \item Second, we concatenate the vertex-represented regions of 2-D GNSS shadows across all the buildings~(Line \ref{line:concatenate_vtx_gnss_shdw}). 

    \item Third, if the $\cno$ value for the current satellite is below a user-specified threshold, then the current satellite is NLOS, so we \emph{intersect} the concatenated 2-D GNSS shadow with our current set-valued position estimate (Line \ref{line:shdw_intersect} and Figure~\ref{fig:ZSM_ex_step_2}); otherwise, the satellite is LOS, so we \emph{subtract} the GNSS shadow from our set-valued position estimate (Line \ref{line:shdw_subtract} and Figure~\ref{fig:ZSM_ex_step_3}).
    We assume the initial AOI (Line \ref{line:init_P}) to contain the true receiver position, and hence, the intersection of GNSS shadow with our current set-valued position estimate is never an empty set.
\end{enumerate}
Finally, after iterating through all buildings and satellites, the algorithm output is a polygon~(2-D polytope) on the ground plane representing the set-valued estimate of receiver positions (potentially with multiple disjoint components) given the current snapshot of GNSS signals  (Line \ref{line:return}).
An illustration of ZSM in simulation is shown later in Figure~\ref{fig:total}.

\subsubsection{Computational Considerations}

We present Algorithm~\ref{alg:zono_shadow_matching} without parallelization for ease of exposition.
However, one can parallelize Algorithm~\ref{alg:zono_shadow_matching} as follows.
When iterating over each $i$\ts{th} building and each $j$\ts{th} satellite, one can create a separate position estimate $P_{i,j}$ for each $(i,j)$ pair, then intersect or subtract each $P_{i,j}$ from the initial position estimate $P$ in Line \ref{line:init_P} by checking if the corresponding building/satellite pair is NLOS or LOS as in Lines \ref{line:if_NLOS}--\ref{line:shdw_subtract}.
Note, we use the unparallelized version of Algorithm~\ref{alg:zono_shadow_matching} in our numerical experiments.

Note that efficient numerical tools, such as MPT \citep{Herceg_2013} and MATLAB's \texttt{polyshape}, exist to perform polytope intersection and subtraction operations in 2-D~(Lines 11-14).
However, to enable leveraging these 2-D polygon tools, we require the constrained zonotope intersection operation as in \eqref{eq:zono_int} to intersect the 3-D building shadow zonotopes with the 2-D ground plane zonotopes.
\section{Experimental Results}\label{sec:experiments}
We experimentally validate our proposed ZSM in simulation. 
Particularly, we choose a simulated platform to conduct exhaustive analysis of our algorithm's performance. 
First, we perform a preliminary experiment to illustrate the utility of our constrained zonotope representation.
Then, we evaluate ZSM in comparison to conventional SM~\citep{Groves_2011,wang2013urban} for two simulation experiments.

Our ZSM implementation is in MATLAB.
We use the open-source CORA toolbox~\citep{althoff2016cora} to represent constrained zonotopes.
We use the \texttt{polyshape} tool to perform set operations on 2-D GNSS shadows.
We utilize received $\cno$ from only GPS; however note that our algorithm is generalizable to other GNSS constellations as well.
As explained earlier in Section~\ref{sec:intro}, we consider an ideal LOS/NLOS classifier in this work, and thus do not account for satellite misclassifications.

\subsection{Preliminary Experiment: Minkowski Sum Evaluation}

To illustrate the utility of using constrained zonotopes, we compare the speed of the Minkowski sum on 1000 randomly-generated polytopes with up to 100 vertices, which we represent as both a standard vertex representation and as constrained zonotopes.
Recall that a convex polytope $P \subset \R^n$ can be represented as the convex hull of a set of vertices $V \subset \R^n$: $P = \convhull{V}$ as in Section~\ref{subsubsec:3-D_map_vertex_rep}.
We use the vertex representation because 3-D urban maps are typically represented as a triangulation of vertices as noted in Section~\ref{subsubsec:3-D_map_vertex_rep}; in other words, this comparison considers an alternative to ZSM wherein we apply Algorithm~\ref{alg:zono_shadow_matching} directly to a 3-D map without converting it to constrained zonotopes.

\begin{figure}[ht]
    \centering
    \includegraphics[width=0.45\textwidth]{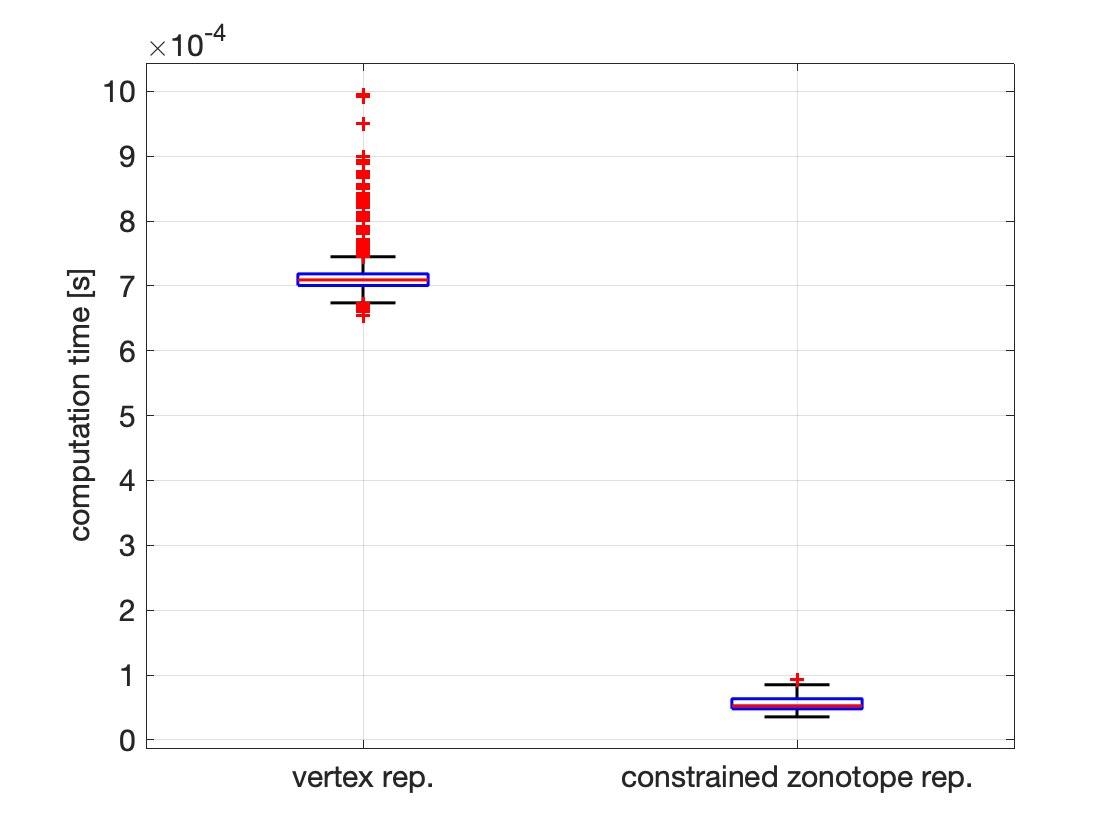}
    \caption{Time comparison for solving Minkowski sums evaluated using a vertex representation (implemented with MPT~\citep{Herceg_2013}) and with a constrained zonotope representation (implemented with CORA \citep{althoff2016cora}).
    The box plot shows the median (red line), 25\ts{th} and 75\ts{th} percentiles (blue blox), interquartile range (blue whiskers), and outliers (red plus signs).
    The zonotope representation is an order of magnitude faster and has a lower standard deviation than vertex representation.}
    \label{fig:mink_sum_comparison}
\end{figure}

We perform the Minkowski sum of both representations with a line segment to mimic the process of computing a building shadow zonotope and compare the average computation time as shown in Figure~\ref{fig:mink_sum_comparison}.
We implement the vertex representation with MPT~\citep{Herceg_2013} and the constrained zonotope representation with CORA~\citep{althoff2016cora}.
This experiment is performed on a laptop computer with an $8$-core $2.4~$GHz CPU and $32~$GB RAM.

The constrained zonotope representation of this operation (executed once) is an order of magnitude faster on average ($71.3~$ms for MPT vs. $5.5~$ms for CORA) and has a smaller standard deviation ($3.3~$ms for MPT vs. $1.1~$ms with fewer outliers for CORA).
Since this Minokwski sum is performed multiple times (specifically, $n\sat \times \prod_{i=1}^{n\bldg} n_i$ times) in Algorithm~\ref{alg:zono_shadow_matching}, we see that the ZSM formulation enables a huge speed increase over using other polytope representations for SM.
We also note that the Minkowski sum for a constrained zonotope with a line segment always results in one additional generator per Figure~\ref{fig:mink_sum_comparison}, whereas it is unclear how many additional vertices may be required for the vertex representation; so it is possible to preallocate memory for the constrained zonotope to further increase performance.

\subsection{Comparison Method and Validation Metrics} \label{sec:validationmetrics}

We validate our proposed ZSM via two simulation experiments: one using a simple 3-D map comprising two buildings and the other using a dense 3-D building map of San Francisco. 
We compare ZSM's performance with that of conventional SM, which is implemented in~\citep{wang2013urban,Groves_2011,Wang_2013}.
All computations are performed on a laptop with a $2$-core $2.5~$GHz CPU and $8~$GB RAM.
In these simulation experiments, we model each building as a constrained zonotope~(i.e., $n_i=1$ in~\eqref{eq:assum_buildings_are_conzonos}.)

\subsubsection{Details of Conventional SM} \label{sec:comparison_setup}

We apply the conventional SM technique presented in~\citep{wang2013urban} and explained earlier in Section~\ref{sec:intro}.
Offline, the method considers a prespecified uniform grid of position candidates and then performs a precomputation step by tracing evenly-distributed azimuth and elevation rays for each position candidate and each building in a 3-D map to compute predicted satellite visibility.
The number of discrete candidates in the uniform grid is determined by its grid size, which is defined as the distance between any two candidates either in cross-street or along-street directions.
Note that the distance between any two position candidates in the along-street direction is the same as in cross-street, thus each grid represents a square.

At runtime~(online computation), conventional SM computes a visibility score at each position candidate by comparing the received $\cno$~(used for LOS/NLOS classification) for each satellite to that of predicted satellite visibility~(from offline computation).
Then, the position candidates with the highest visibility scores~(can be more than one) are identified as the most likely receiver positions.
The conventional method also computes the weighted empirical covariance~\citep{Martens_2003} by analyzing the position candidates and their associated weights, which are based on the normalized visibility scores.

To analyze ZSM performance for a given initial AOI, we precompute the visibility map of conventional SM for various grid sizes. 
As mentioned earlier in Section~\ref{sec:intro}, a finer discretization is attained with a lower grid size, thus increasing the position accuracy of the conventional SM technique, but at the expense of higher computation cost~(both offline and online).

\subsubsection{Validation Metrics}

We apply four validation metrics:
\begin{enumerate}
    \item Offline computation load: 
    For ZSM, this is the time required for converting buildings from vertices to constrained zonotopes.
    For conventional SM, this is the time required for precomputing the visibility map at each position candidate, and thus the predicted satellite visibility.
    
    \item Online computation load:
    For ZSM, this is the time required to run Algorithm~\ref{alg:zono_shadow_matching}.
    For conventional SM, this is the time required to compute the visibility scores at all position candidates and compute the most likely position candidates~(based on highest visibility scores) and the weighted empirical covariance.
    
    \item Point-valued estimation error in cross-street and along-street directions:
    For ZSM, this is the error in cross-street and along-street directions between the true receiver location and centroid of the final set-valued position estimate (multiple centroid values obtained, when disjoint components are present).
    For conventional SM, this is the error in cross-street and along-street directions between the most likely position candidates~(highest visibility score) and the true receiver location.
    
    \item Bounds in cross-street and along-street directions:
    For ZSM, this is the width of the bounding box that encloses the set-valued position estimate. Note that we report the width of individual bounding boxes, when disjoint components are present in the final set-valued position estimate.
    For conventional SM, this is twice the 3$\sigma$ (three standard deviations) bound of the visibility-based weighted empirical covariance in the cross-street and along-street directions.
    In particular, we evaluate the weighted empirical covariance across all the position candidates based on their visibility scores. 
\end{enumerate}

\subsection{Simulated Experiment \#1: Simple 3-D Building Map} \label{sec:zsm_vs_csm}

We first describe a simulation experiment that considers a simple 3-D map comprising two urban buildings. 
We use the real-world GPS data collected from Android phone and then, emulate the received $\cno$ values with simulated NLOS effects.

\subsubsection{GPS Dataset and NLOS Emulation} \label{sec:first_nlos_emulation}

We use the openly-available Google Android dataset~\citep{Fu_2020}, with the data collection setup shown in Figure~\ref{fig:GoogleAndroid}. 
To allow control over LOS/NLOS signals for each satellite, we use the data collected in open-sky conditions via a Pixel 4 XL Modded smartphone, which logs the GPS data, and a high-fidelity GNSS-RTK/IMU setup, which serves as a reference ground truth.

To design an ideal LOS/NLOS classifier, we emulate NLOS effects as follows.
First, we overlay the test surroundings with simulated 3-D buildings so as to mimic an urban setting. 
We then choose a time when nine satellites are LOS, as judged by $\cno$ values above 38 dB-Hz, which is representative of open-sky conditions based on the empirical study conducted in~\citep{Kuusniemi_2004,hetet2000signal}.
We induce simulated NLOS effects in a GPS satellite when the LOS vector between its location and the true receiver position is obstructed by any simulated 3-D building.
In particular, we attenuate the received $\cno$ values for these identified NLOS satellites to be below the 38 dB-Hz threshold.
Given that we are considering an ideal LOS/NLOS classifier that is based on binary thresholding against 38 dB-Hz, we do not account for the emulation and inclusion of multipath effects in this work.

\subsubsection{Experiment Setup}
\begin{figure}[H]
	\setlength{\belowcaptionskip}{-4pt}
	\centering
	\begin{subfigure}[b]{0.49\textwidth}
	    \centering
		\includegraphics[scale=0.35]{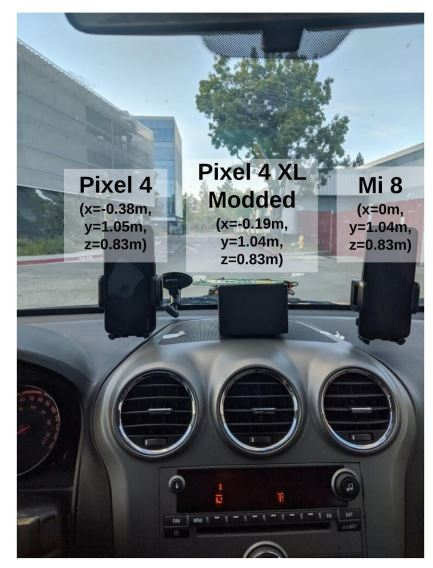}
	\caption{Pixel 4 XL Modded smartphone used for data collection}
	\label{fig:GoogleAndroid}
	\end{subfigure}
	\hspace{-5mm}
	\begin{subfigure}[b]{0.49\textwidth}
	    \centering
	    \includegraphics[scale=0.25]{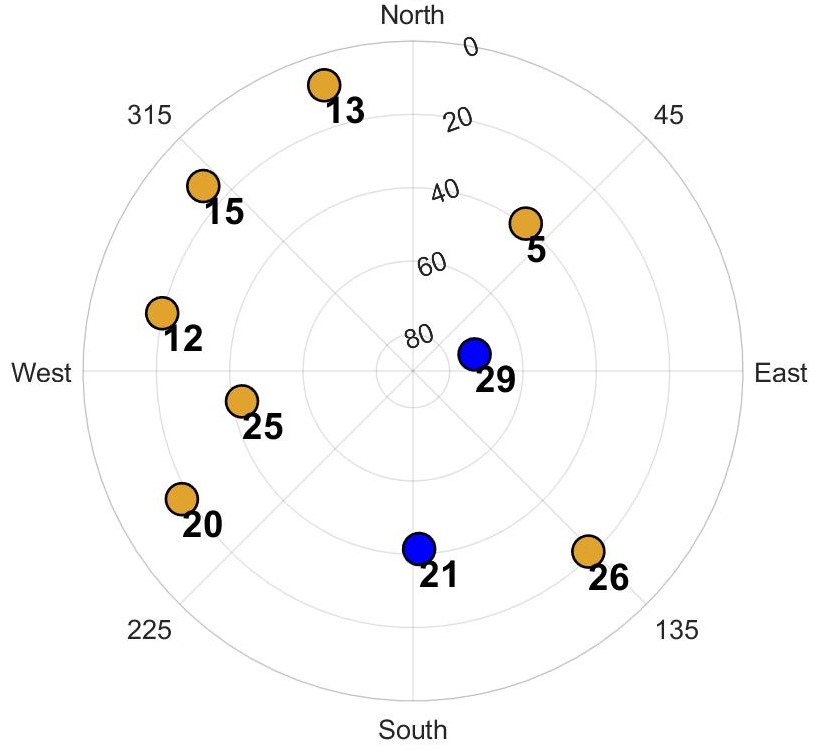}
	\caption{GPS satellite skyplot}
	\label{fig:skyplot}
	\end{subfigure}
	\caption{Subfigure (a) is the experiment test platform utilized to validate the proposed ZSM via publicly-available Google Android datasets in~\citep{Fu_2020}.
	Subfigure (b) shows the skyplot for a particular receiver location with nine visible GPS satellites.
	The GPS satellites are represented by circles with LOS satellites in blue and NLOS~(simulated effects) in dark yellow.}
	\label{fig:dataset}
\end{figure}

We set up the experiment as follows, also shown in Figure~\ref{fig:ZSM_ex_overview}.
For the urban 3-D map, we consider two buildings.
The receiver 2-D position is initialized at $(0,0)~$m in local map coordinates, and the initial AOI is chosen as the entire length and width of a street between the two buildings, shown later in Figure~\ref{fig:ZSM_ex_step_1}.
We observe a total of nine visible GPS satellites whose skyplot is shown in Figure~\ref{fig:skyplot}.  
As mentioned earlier, we induce simulated errors in received $\cno$ to represent LOS/NLOS characteristics with respect to the true receiver position and 3-D map.
In Figure~\ref{fig:skyplot}, the NLOS satellites are indicated by dark yellow circle markers whereas LOS ones are blue. 
For conventional SM, we consider three grid sizes with the distances between the discrete candidates in cross-street and along-street directions as follows: $15~$m, $10~$m, and $5~$m.
The number of position candidates for the grid sizes of $15~$m, $10~$m, and $5~$m, are $63$, $124$, and $427$ respectively.

\subsubsection{Results and Discussion}

\begin{figure}[H]
	\centering
	\begin{subfigure}[b]{0.49\textwidth}
		\includegraphics[width=0.98\textwidth]{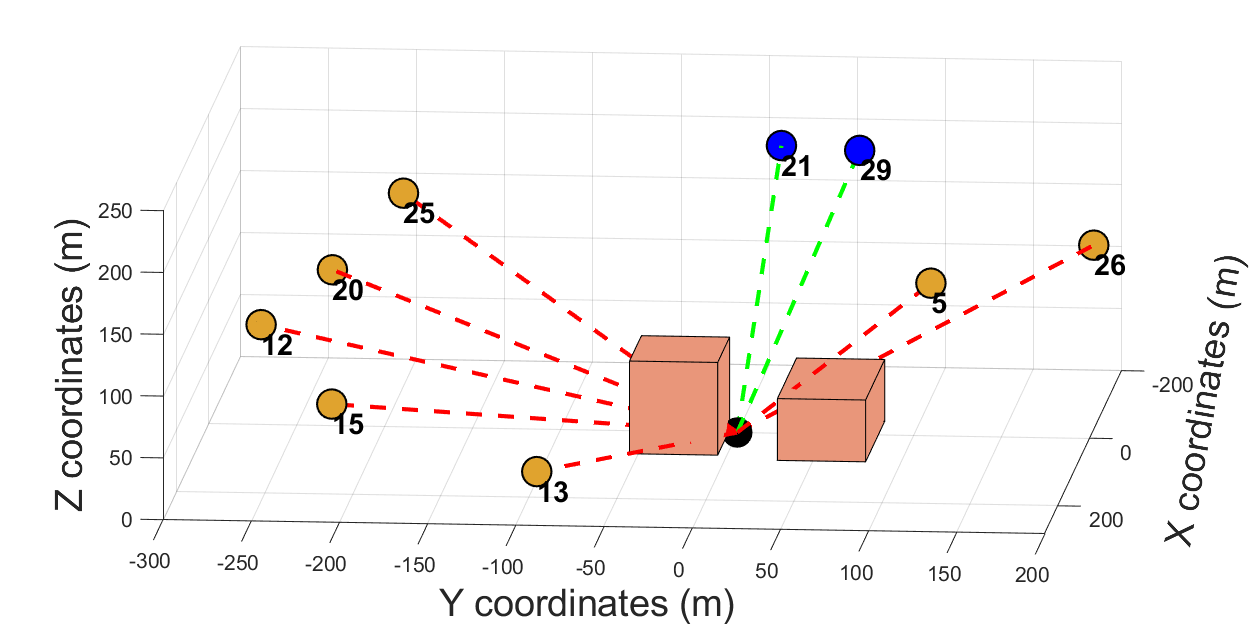}
		\caption{Simulated setup comprising nine satellites}
		\label{fig:ZSM_ex_overview}
	\end{subfigure}	
	\hspace{2mm}
	\begin{subfigure}[b]{0.49\textwidth}
		\includegraphics[width=0.8\textwidth]{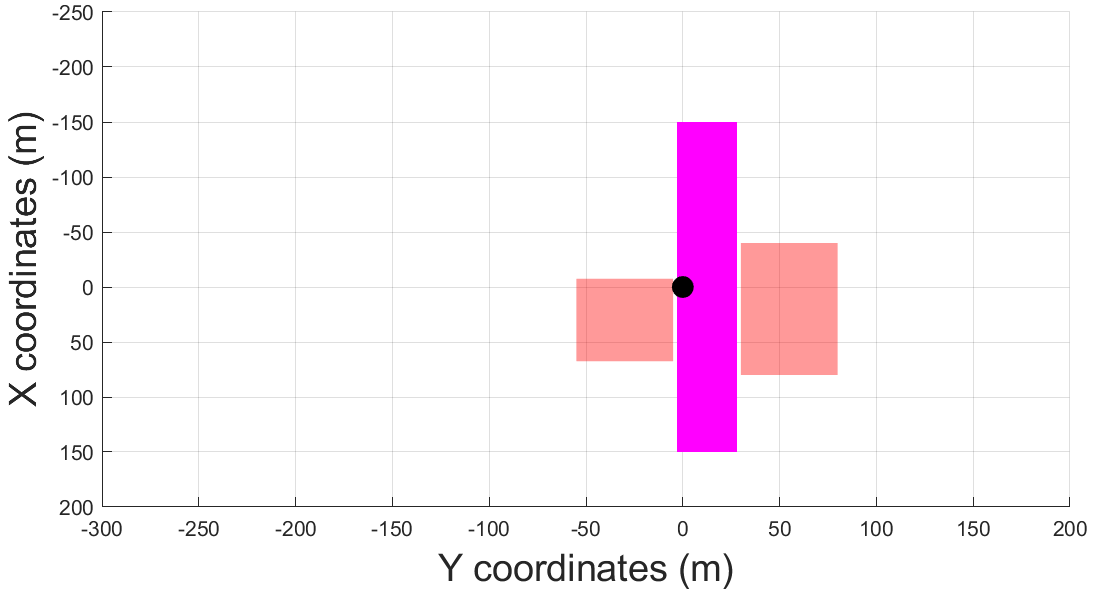}
		\caption{Top view: Initial polygon receiver position (magenta)}
		\label{fig:ZSM_ex_step_1}
	\end{subfigure}
	\vfill
	\vspace{5mm}
	\begin{subfigure}[b]{0.49\textwidth}
		\includegraphics[width=0.98\textwidth]{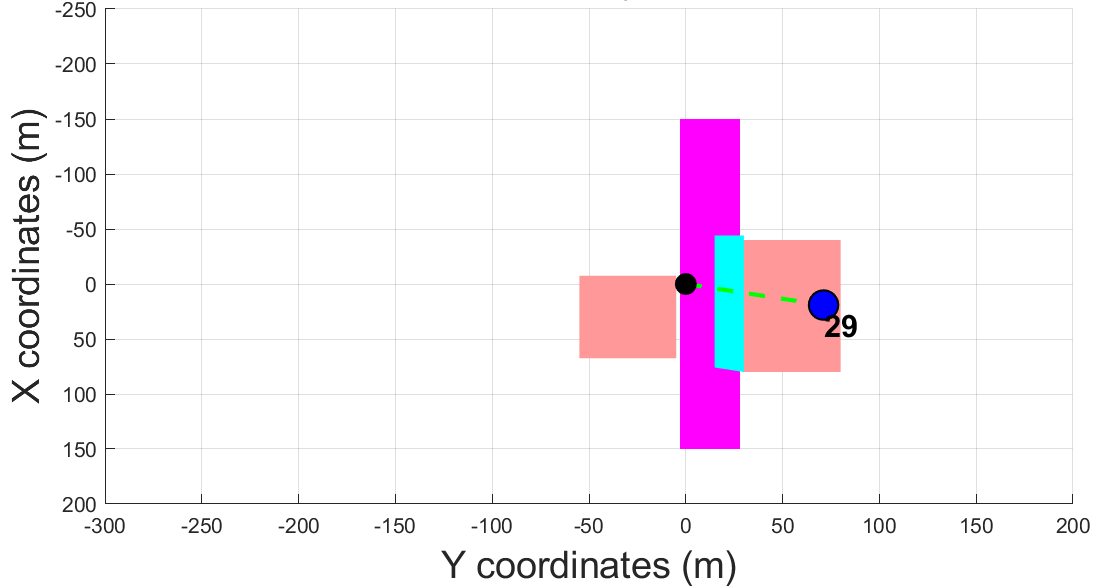}
		\caption{Top view: Set difference for PRN $29$}
		\label{fig:ZSM_ex_step_2}
	\end{subfigure}
	\hspace{1mm}
	\begin{subfigure}[b]{0.49\textwidth}
		\includegraphics[width=0.98\textwidth]{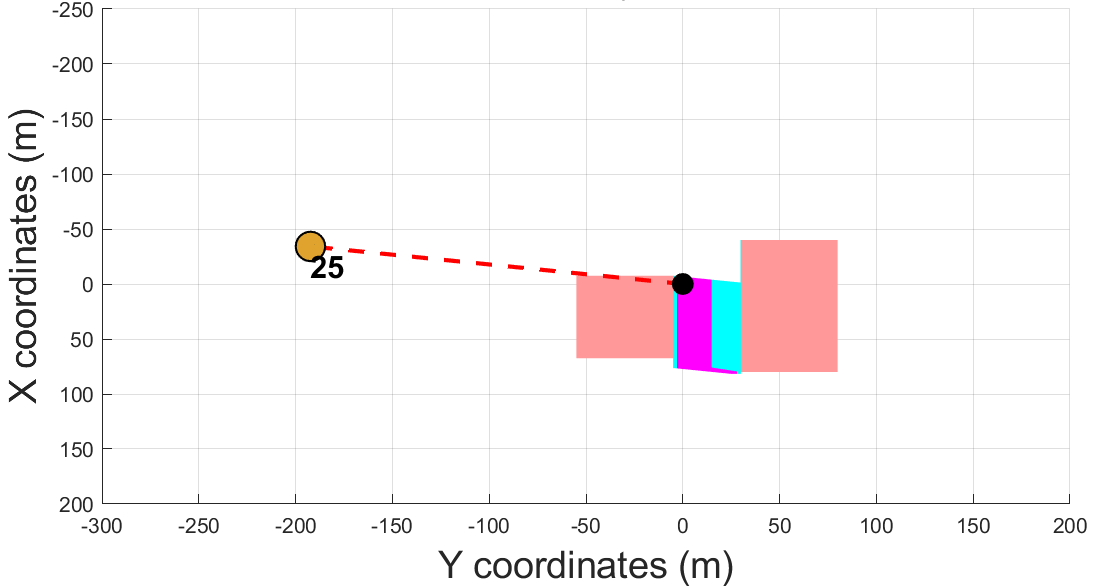}
		\caption{Top view: Set intersection for PRN $25$}
		\label{fig:ZSM_ex_step_3}
	\end{subfigure}	
	\vfill
	\vspace{5mm}
	\begin{subfigure}[b]{0.49\textwidth}
		\includegraphics[width=0.98\textwidth]{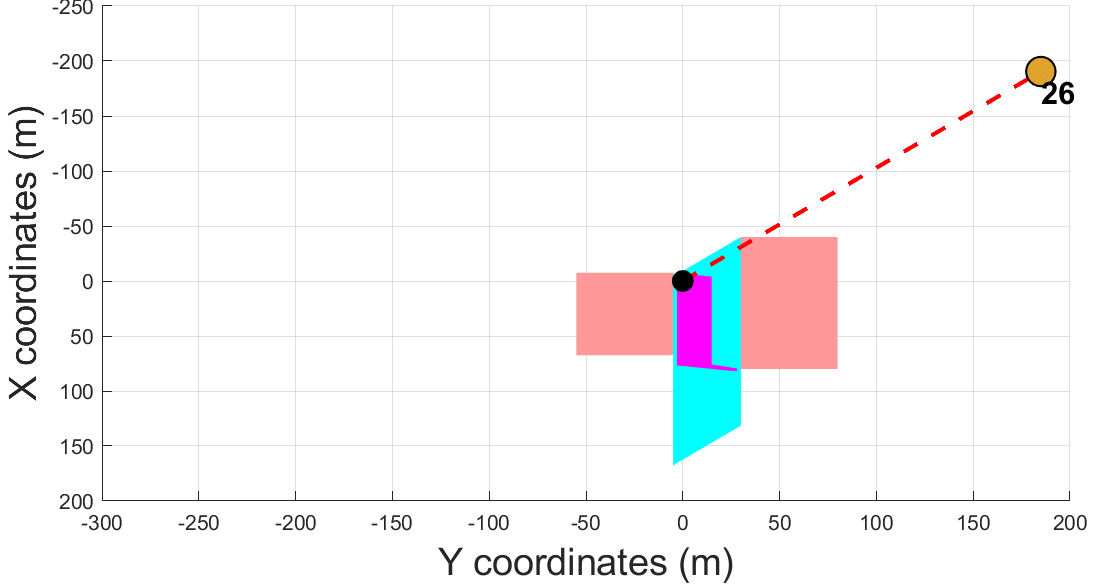}
		\caption{Top view: Set intersection for PRN $26$}
		\label{fig:ZSM_ex_step_4}
	\end{subfigure}
	\hspace{1mm}
	\begin{subfigure}[b]{0.49\textwidth}
		\includegraphics[width=0.98\textwidth]{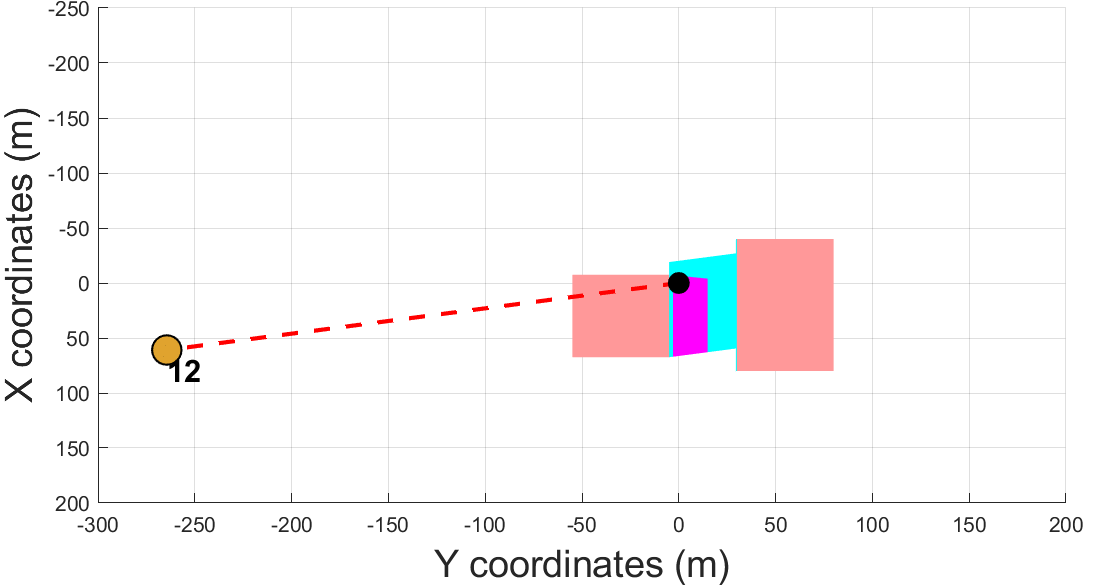}
		\caption{Top view: Set intersection for PRN $12$}
		\label{fig:ZSM_ex_step_5}
	\end{subfigure}	
	\caption{ZSM experiment setup and procedure; (a)~shows the simulated setup with two urban buildings and nine GPS satellites, with the true receiver position as a black circle, the LOS satellites as a blue circles with LOS signals as a green dashed line, and the NLOS satellites as dark yellow circles with NLOS signals as red dashed line; (b)~shows a top-down view with the initial AOI in magenta; (c)-(f) show the GNSS shadow for each satellite/building pair in cyan and the resulting receiver position estimate after set intersection/difference in magenta.
	While we only illustrate the procedure for PRNS 29, 25, 26, and 12, the final result from all nine satellites is shown in Figure~\ref{fig:performance_analysis}.
	Note that the satellites are plotted near the receiver for visualization only; the shadow directions as in \eqref{eq:shadow_dir} are computed using the satellites' actual positions.
	Given the received $\cno$ and 3-D map, we demonstrate that the proposed ZSM computes a set-valued receiver position estimate.}
	\label{fig:total}
\end{figure}

\begin{figure}[H]
	\centering
	\begin{subfigure}[b]{0.49\textwidth}
		\includegraphics[width=0.98\textwidth]{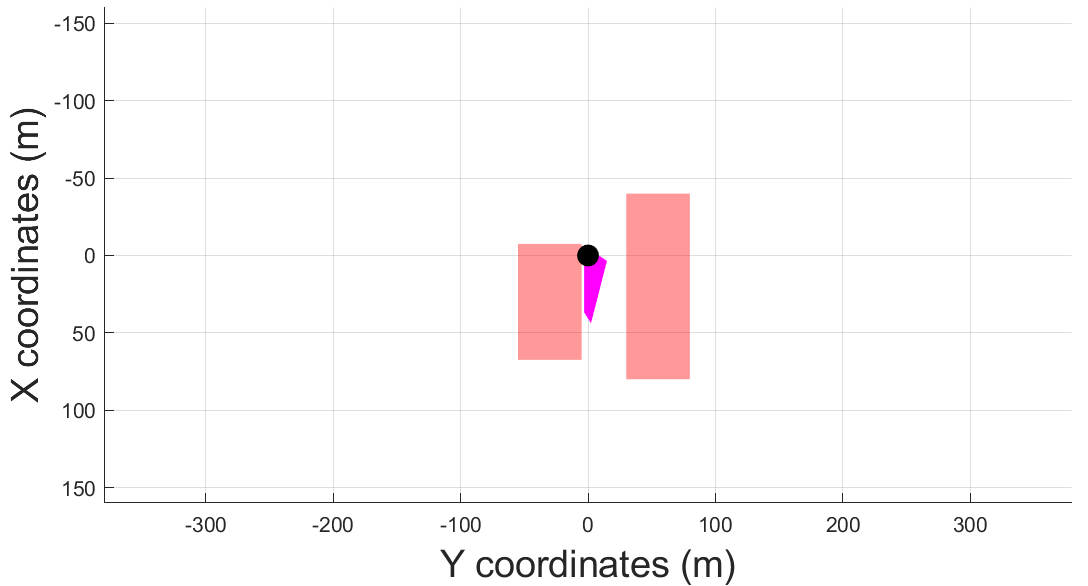}
		\caption{Proposed Zonotope Shadow Matching (ZSM)}
		\label{fig:proposedZSM_first}
	\end{subfigure}	
	\hspace{2mm}
	\begin{subfigure}[b]{0.49\textwidth}
		\includegraphics[width=0.98\textwidth]{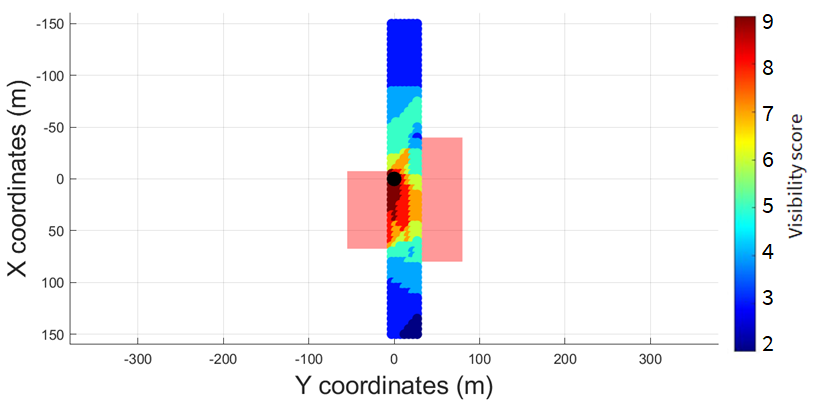}
		\caption{Conventional SM with grid size $5~$m}
		\label{fig:conventionalSM_5}
	\end{subfigure}
	\begin{subfigure}[b]{0.49\textwidth}
		\includegraphics[width=0.98\textwidth]{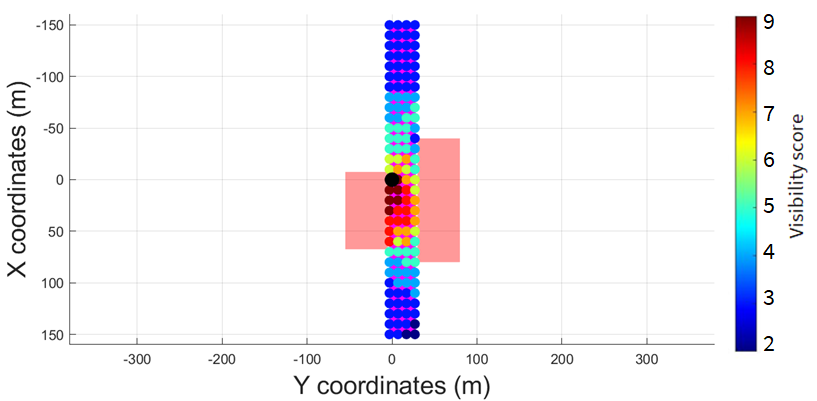}
		\caption{Conventional SM with grid size $10~$m}
		\label{fig:conventionalSM_10}
	\end{subfigure}
	\begin{subfigure}[b]{0.49\textwidth}
		\includegraphics[width=0.98\textwidth]{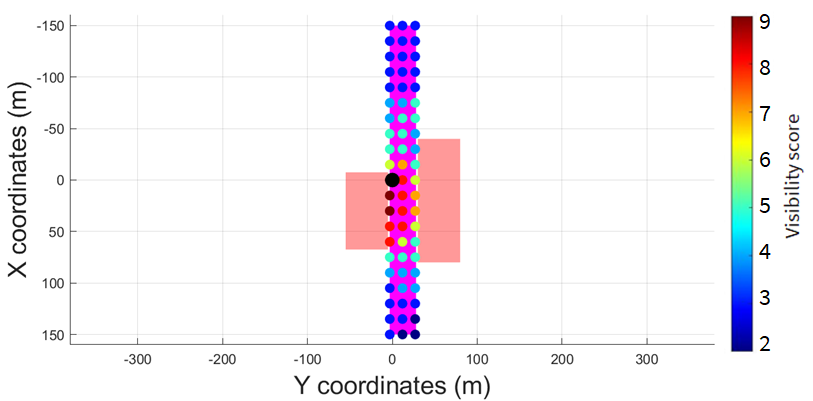}
		\caption{Conventional SM with grid size $15~$m}
		\label{fig:conventionalSM_15}
	\end{subfigure}	
	\caption{Performance analysis of the proposed ZSM vs. conventional SM for a simple 3-D building map; (a)~shows the ZSM receiver position estimate in magenta; (b),~(c) and (d)~shows the conventional SM with grid sizes of $5~$m, $10~$m and $15~$m, respectively, where the position candidates are color-coded using jet colormap with blue indicating a low visibility score of $2$ and red representing a high value of $9$~(number of available GPS satellites).
	For a lower online computation time of $0.39~$s, ZSM demonstrates a comparable point-valued estimation error and a smaller width than the densest conventional SM case in both cross-street and along-street directions.}
	\label{fig:performance_analysis}
\end{figure}

The first few steps~(top-down view) of Algorithm~\ref{alg:zono_shadow_matching} are shown in Figures~\ref{fig:ZSM_ex_step_1}-\ref{fig:ZSM_ex_step_5}.
In particular, we illustrate the satellites, buildings, initial AOI, GNSS shadows, and intersection/subtraction procedure.
The full results are summarized in Table~\ref{table:first_stats} and the final localization results for proposed ZSM and conventional SM are illustrated in Figures~\ref{fig:proposedZSM_first}-\ref{fig:conventionalSM_5}.
For each grid size of conventional SM reported in Table~\ref{table:first_stats}, we report the top three most-likely position candidates~(ranked based on the highest visibility score and the least point-valued estimation error with respect to ground truth). 

The results from the three cases of the conventional SM technique are used for validating the improved performance of our proposed ZSM algorithm.
We see a comparable point-valued estimation accuracy of ZSM, i.e., $3.46~$m and $16.05~$m in cross-street and along-street directions, respectively, as that of conventional SM with grid sizes of $5~$m, $10~$m and $15~$m, whose accuracies are reported in Table~\ref{table:first_stats}.
Furthermore, unlike all the conventional SM cases, we show that ZSM achieves a set-valued position estimate with an accuracy in width of only $17.87~$m in cross-street direction and $50.11~$m in along-street direction. 
We also find that the offline and online computations of ZSM only incurs an average runtime of $1.7 \times 10^{-4}~$s and $0.39~$s, respectively.
Importantly, without the need for gridding, the ZSM result produces this point-valued positioning accuracy while returning a set-valued estimate, which contains the true receiver position.
We anticipate that the result might be more accurate if we leverage more GPS satellites and buildings that provide additional geometry constraints in the shadow matching, thus reducing the centroid error and bounds of the disjoint component~(ambiguous mode) containing the true receiver location, if any. 
This forms the premise of our next simulation experiment.

\begin{table}[H]
\centering 
\caption{Performance comparison results for Zonotope Shadow Matching (ZSM) vs conventional SM on an simulation experiment with two buildings seen in Figure~\ref{fig:ZSM_ex_overview} and nine GPS satellites seen in Figure~\ref{fig:skyplot}.}
\begin{tabular}{|c|c|c|c|c|c|}
\hline
\multicolumn{2}{|c|}{\multirow{2}{*}{\textbf{Algorithm}}} & \multirow{2}{*}{\textbf{\begin{tabular}[c]{@{}c@{}}Error w.r.t true location (m)\\ {[}Cross-street, Along-street{]}\end{tabular}}} & \multirow{2}{*}{\textbf{\begin{tabular}[c]{@{}c@{}}Bound (m)\\ {[}Cross-street, Along-street{]}\end{tabular}}} & \multicolumn{2}{c|}{\textbf{\begin{tabular}[c]{@{}c@{}}Avg. computation load \\ across 100 runs (s)\end{tabular}}} \\ \cline{5-6} 
\multicolumn{2}{|c|}{} &  &  & \textbf{Offline} & \textbf{Online} \\ \hline
\multicolumn{2}{|c|}{\textbf{Proposed ZSM}} & {[}3.46, 16.05{]} & {[}\textbf{17.87, 50.11}{]} & \textbf{1.7e-4} & \textbf{0.39} \\ \hline
\multirow{3}{*}{\textbf{\begin{tabular}[c]{@{}c@{}}Conventional \\ SM with \\ grid sizes\end{tabular}}} & \textbf{5 m} & \begin{tabular}[c]{@{}c@{}}{[}3.00, 5.00{]}\\ {[}\textbf{2.00, 0.00}{]}\\ {[}3.00, 0.00{]}\end{tabular} & {[}59.58, 447.48{]} & 1524.41 & 1.66 \\ \cline{2-6} 
 & \textbf{10 m} & \begin{tabular}[c]{@{}c@{}}{[}3.00, 0.00{]}\\ {[}7.00, 0.00{]}\\ {[}3.00, 10.00{]}\end{tabular} & {[}66.72, 455.82{]} & 405.33 & 0.08 \\ \cline{2-6} 
 & \textbf{15 m} & \begin{tabular}[c]{@{}c@{}}{[}3.00, 0.00{]}\\ {[}3.00, 15.00{]}\\ {[}3.00, 30.00{]}\end{tabular} & {[}73.54, 463.10{]} & 296.32 & 0.04 \\ \hline
\end{tabular}
\label{table:first_stats}
\end{table}

\subsection{Simulated Experiment \#2: Using 3-D Building Map of San Francisco}

We perform our next simulation experiment using a publicly-available 3-D building map of San Francisco. 
For this dense 3-D urban map, we validate our algorithm's performance as compared to that of the conventional SM technique using emulated GPS data. 
We also perform sensitivity analysis of proposed ZSM by varying the number of buildings being considered. 

\subsubsection{GPS Dataset and NLOS Emulation} \label{sec:GPSdatasetandNLOSemulation}
\begin{figure}[ht]
    \centering
    \includegraphics[width=0.9\textwidth]{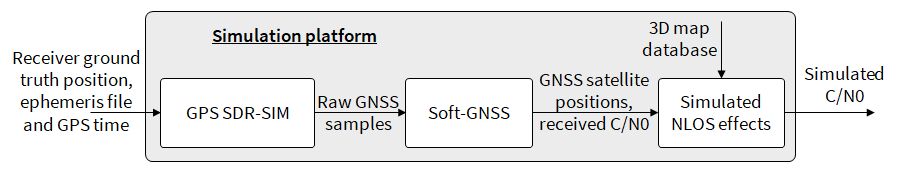}
    \caption{Our GPS software simulation pipeline uses two publicly-available software: GPS-SDR-SIM~\citep{gpssdrsim} and SoftGNSS~\citep{softGNSS}. 
    For a given true receiver location, we simulate the $\cno$ values and thereafter, induce NLOS effects using 3-D building map of San Francisco. }
    \label{fig:software_simulator}
\end{figure}

We generate GPS data using the simulation pipeline shown in Fig.~\ref{fig:software_simulator}. 
Considering a desired static position~(true receiver location), ephemeris file and start time as inputs, we utilize a C++ language-based software-defined GPS simulator known as GPS-SIM-SDR~\citep{gpssdrsim,Bhamidipati_2019} to simulate the raw GPS samples, which are indicative of open-sky conditions. 
Later, we perform acquisition and tracking using a MATLAB-based software-defined radio known as SoftGNSS~\citep{softGNSS} to generate received $\cno$ values. 
Similar to NLOS emulation described in Section~\ref{sec:first_nlos_emulation}, we induce NLOS effects in $\cno$ values based on the simulated 3-D buildings in San Francisco and true receiver location.  

\subsubsection{Experiment Setup}
\begin{figure}[ht]
	\setlength{\belowcaptionskip}{-4pt}
	\centering
	\begin{subfigure}[b]{0.45\textwidth}
	    \centering
		\includegraphics[width=0.98\textwidth]{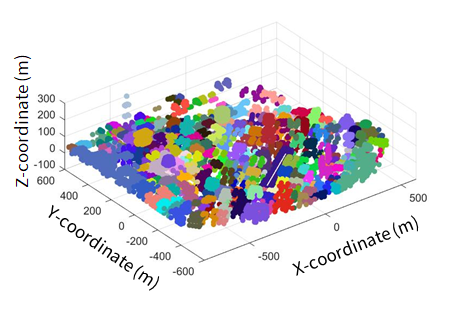}
	\caption{Vertex representation}
	\label{fig:VertexRepresentation}
	\end{subfigure}
	\hspace{5mm}
	\begin{subfigure}[b]{0.45\textwidth}
	    \centering
	    \includegraphics[width=0.98\textwidth]{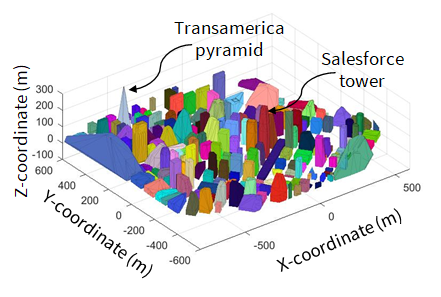}
	\caption{Constrained zonotope representation}
	\label{fig:ConZonoRepresentation}
	\end{subfigure}
	\caption{Preprocessing 3-D building map of San Francisco; (a)~shows the vertex representation, wherein the vertices of each building are separately color-coded; (b)~represents each building as a constrained zonotope, which is given as input to our proposed ZSM algorithm.}
\end{figure}

We utilize a publicly-available 3-D building map of San Francisco~\citep{gyorgy2018sfmap}. 
As explained in~\eqref{eq:building_are_triangles} of Section~\ref{subsubsec:3-D_map_vertex_rep}, standard 3-D building maps are represented as union of triangles, with each triangle comprising three vertices. 
An illustration of the vertex representation for the 3-D building map of San Francisco is shown in Figure~\ref{fig:VertexRepresentation}, wherein we isolate the vertices of each building using the theory of connected graphs~\citep{K_nig_1990}, and color code them independently. 
We then convert the entire 3-D map from vertex representation to constrained zonotope representation, which can be visualized in Figure~\ref{fig:ConZonoRepresentation}. 
Note that, the prominent landmarks in San Francisco, such as Transamerica pyramid and Salesforce tower, can be easily identified in Figure~\ref{fig:ConZonoRepresentation}. 
The storage space required for building constrained zonotopes is quite less. 
For instance, the .mat file size for $8$ buildings discussed in Section~\ref{subsec:sensitivityanalysis} is $5$~KB while for $14$ buildings is $8$~KB and for $20$ buildings is $11$~KB. 
Similarly, for the entire San Francisco map (shown in Figure~\ref{fig:ConZonoRepresentation}) is $345$~KB. While this is acceptable for many practical applications, based on the user platform specifications, if further reduction in file size is required, one can employ alternate techniques such as storing a reduced order representation of building constrained zonotopes or storing a smaller section of the map at the original order of the building constrained zonotopes.
Note that the experiments conducted in this subsection are fully simulated, and therefore, inaccuracies of this 3-D building map compared to that of the San Francisco city are not relevant. 
This is because we simulate the true receiver location and classify the satellites as LOS/NLOS relative to this open-source 3-D building map.

\begin{figure}[H]
	\setlength{\belowcaptionskip}{-4pt}
	\centering
	\begin{subfigure}[b]{0.33\textwidth}
	    \centering
	    \includegraphics[width=0.98\textwidth]{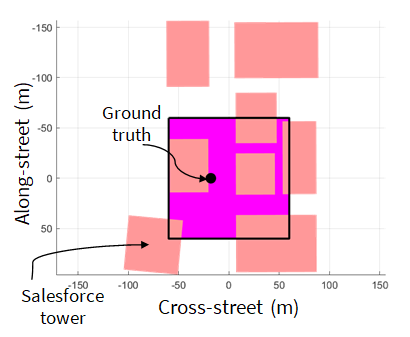}
	\caption{Initial AOI}
	\label{fig:second_initialAOI}
	\end{subfigure}
	\begin{subfigure}[b]{0.33\textwidth}
	    \centering
	    \includegraphics[width=0.98\textwidth]{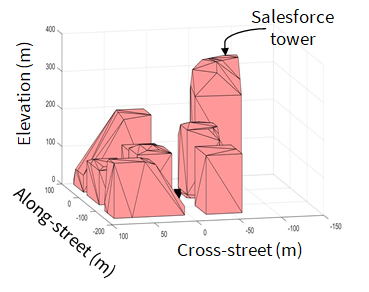}
	\caption{A relevant section from 3-D San Francisco map}
	\label{fig:second_3Dsetup}
	\end{subfigure}
	\begin{subfigure}[b]{0.33\textwidth}
	    \centering
		\includegraphics[width=0.98\textwidth]{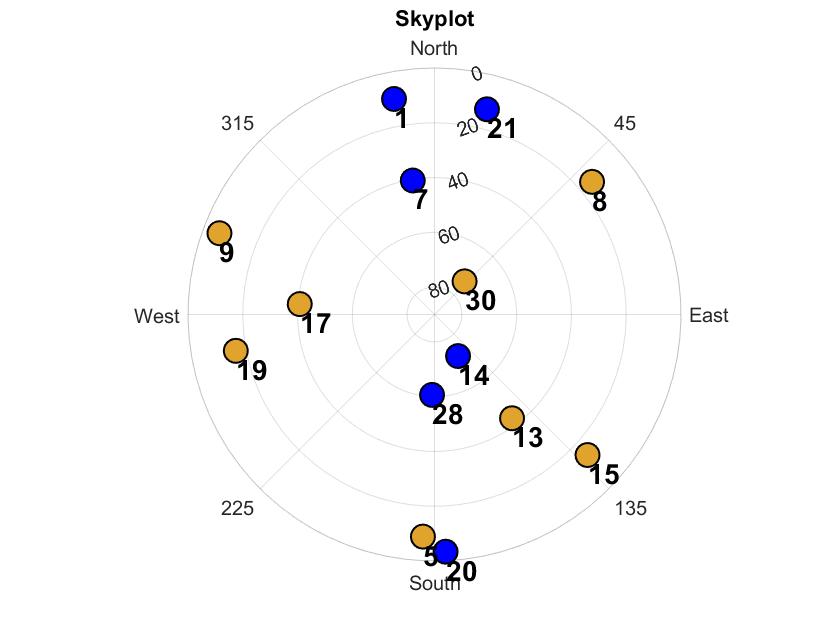}
	\caption{GPS satellite skyplot}
	\label{fig:second_skyplot}
	\end{subfigure}
	\caption{Our second simulation experiment using 3-D map of San Francisco. Subfigure (a) shows the initial AOI of size $120~$m$\times 120~$m in a top-down view. 
	Subfigure (b) shows the relevant section of 3-D map near the Salesforce tower, which is a prominent landmark in San Francisco.
	Subfigure (c) shows the skyplot with respect to true receiver location with fourteen visible GPS satellites. 
	The GPS satellites are represented by circles with LOS satellites in blue and NLOS~(simulated effects) in dark yellow.}
\end{figure}
The true receiver position is initialized at $(0,-18)~$m in local map coordinates. 
The initial AOI is chosen to lie within a size of $120~$m$\times 120~$m while excluding the regions that lie within the building footprints. 
In particular, we choose the size of AOI in a heuristic manner by penalizing the position solution and uncertainty bounds~($3\sigma$) estimated by the MATLAB-based SoftGNSS~\citep{softGNSS}, which was explained earlier in Section~\ref{sec:GPSdatasetandNLOSemulation}.
A top-down view of this illustration in shown in Figure~\ref{fig:second_initialAOI} and a relevant section of 3-D San Francisco map comprising eight buildings~(of which one of them is the Salesforce tower) is shown in Figure~\ref{fig:second_3Dsetup}.
We simulate fourteen GPS satellites whose skyplot is shown in Figure~\ref{fig:second_skyplot}, with LOS satellites indicated by blue circle markers and NLOS by dark yellow. 
For comparison with the conventional SM technique~(explained earlier in Section~\ref{sec:validationmetrics}), we consider four grid sizes of $30~$m, $10~$m, $5~$m and $3~$m.
The number of position candidates for the grid sizes of $30~$m, $10~$m, $5~$m and $3~$m, are $16$, $97$, $346$ and $902$, respectively.

\subsubsection{Comparison Results with Conventional SM} \label{subsec:second_simulation}
\begin{figure}[H]
	\centering
	\begin{subfigure}[b]{0.48\textwidth}
		\includegraphics[width=0.98\textwidth]{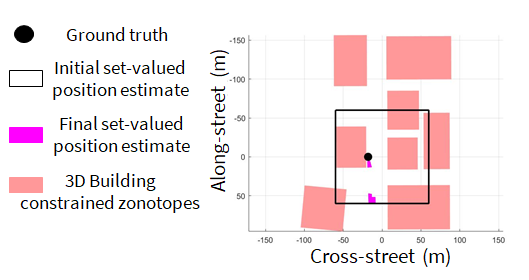}
		\caption{ZSM with all available $14$ satellites}
		\label{fig:second_proposedZSM}
	\end{subfigure}	
	\hspace{5mm}
	\begin{subfigure}[b]{0.32\textwidth}
		\includegraphics[width=0.98\textwidth]{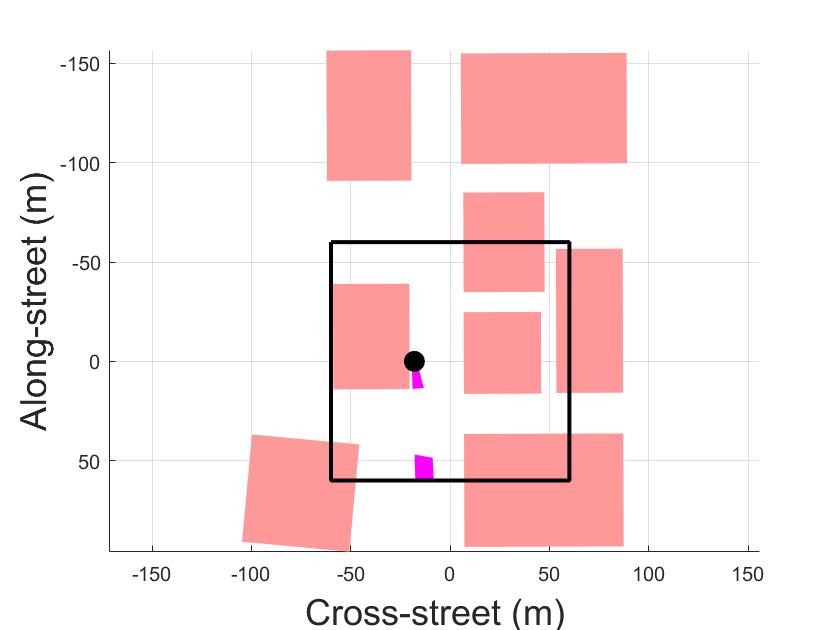}
		\caption{ZSM with $10$ random satellites among $14$}
		\label{fig:second_proposedZSM_10sats}
	\end{subfigure}
	\begin{subfigure}[b]{0.32\textwidth}
		\includegraphics[width=0.98\textwidth]{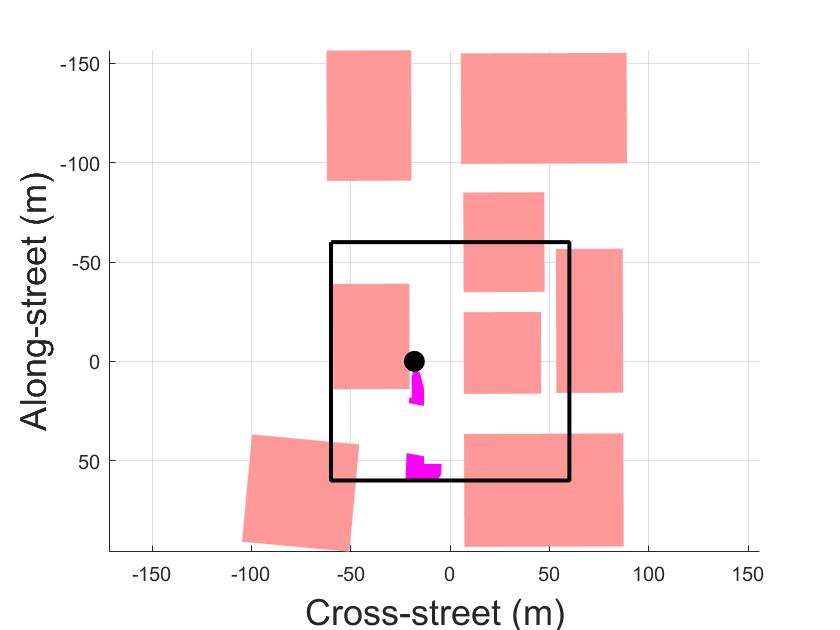}
		\caption{ZSM with $8$ random satellites among $14$}
		\label{fig:second_proposedZSM_8sats}
	\end{subfigure}
	\hspace{10mm}
	\begin{subfigure}[b]{0.32\textwidth}
		\includegraphics[width=0.98\textwidth]{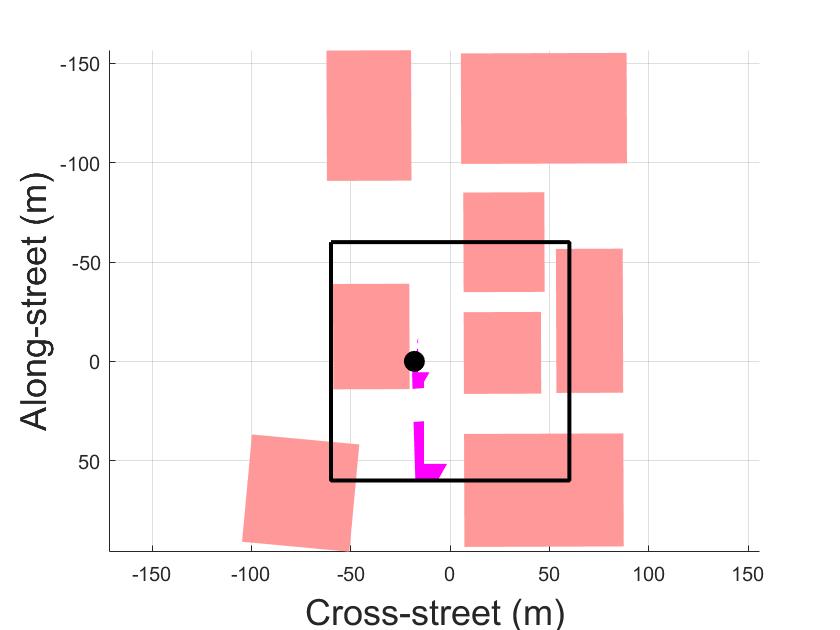}
		\caption{ZSM with satellites above elevation of $20^\circ$}
		\label{fig:second_proposedZSM_20elev}
	\end{subfigure}
	\caption{Performance analysis of the proposed ZSM as the number of GPS satellites are varied, wherein the final set-valued receiver position estimate is indicated in magenta, the building footprints in red, and the initial AOI by a black box; 
	Our proposed ZSM algorithm leveraging all the available $14$ satellites successfully estimates not only a high point-valued position accuracy of $1.2~$m and $8.1~$m in cross-street and along-street directions, respectively, but also small position bounds of $5.8~$m and $8.1~$m in cross-street and along-street directions, respectively.
	We observe that the mode containing the true receiver location remains majorly unaffected as the number of GPS satellites decrease from $14$~(all available ones) to a subset of GPS satellites, i.e., $10$, $8$ and $6$~(elevation $>20^{\circ}$).}
\end{figure}

\begin{figure}[H]
	\centering
	\begin{subfigure}[b]{0.43\textwidth}
		\includegraphics[width=0.98\textwidth]{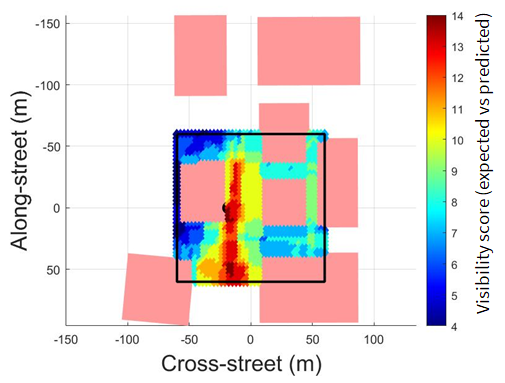}
		\caption{Conventional SM with $3~$m grid size}
		\label{fig:second_conventionalSM_3m}
	\end{subfigure}
	\begin{subfigure}[b]{0.43\textwidth}
		\includegraphics[width=0.98\textwidth]{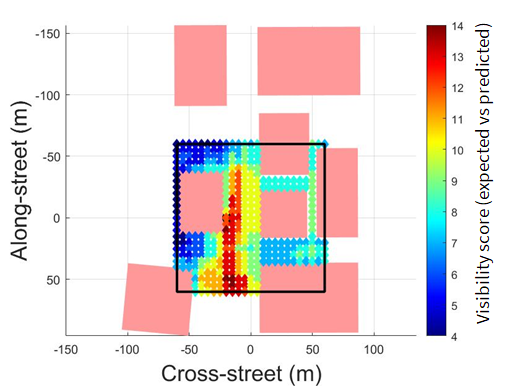}
		\caption{Conventional SM with $5~$m grid size}
		\label{fig:second_conventionalSM_5m}
	\end{subfigure}
	\hfill
	\begin{subfigure}[b]{0.43\textwidth}
		\includegraphics[width=0.98\textwidth]{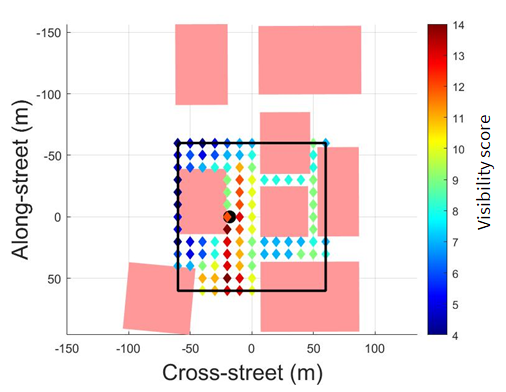}
		\caption{Conventional SM with $10~$m grid size}
		\label{fig:second_conventionalSM_10m}
	\end{subfigure}
	\begin{subfigure}[b]{0.43\textwidth}
		\includegraphics[width=0.98\textwidth]{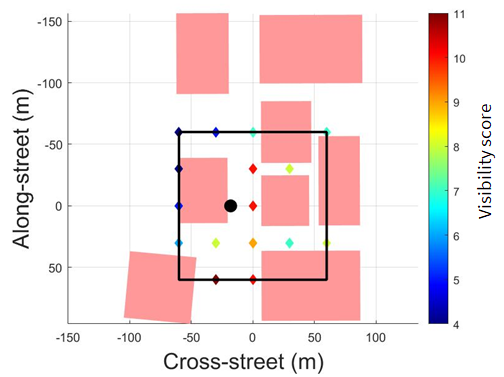}
		\caption{Conventional SM with $30~$m grid size}
		\label{fig:second_conventionalSM_30m}
	\end{subfigure}
	\caption{Performance analysis of the conventional SM with grid size of $3~$m, $5~$m, $10~$m, and $30~$m, respectively, which are color-coded using jet colormap with blue indicating a low visibility score of $0$~(no GPS satellites match) and red indicating a high value of $14$~(all the available ones match) for a)-c)
	For d), given the larger grid size, there is no grid point where predicted vs. expected visibilities match for all GPS satellites, thus red in its colorbar indicating a high value of $11$. 
	We observe that, as the grid resolution increases or grid size decreases, the most likely candidate positions from conventional SM closely resemble the centroid of two disjoint sets estimated via proposed ZSM in Figure~\ref{fig:second_proposedZSM}.
	Also, we observe that the uncertainty bounds estimated using our conventional SM exceeds $100~$m for all grid sizes, thus validating the significance of proposed ZSM that can compute exact bounds without requiring to discretize.
}
\end{figure}

able~\ref{table:second_stats} reports comparison statistics between the proposed ZSM with different satellite subsets and conventional SM with varying grid sizes in terms of offline and online computation load, point-valued estimation error with respect to true receiver location, and the associated position bounds. 
A visualization of our ZSM with all available $14$ satellites, a random subset of $10$ satellites, a random subset of $8$ satellites and satellites with elevation~$>20^{\circ}$ are shown in Figures~\ref{fig:second_proposedZSM},~\ref{fig:second_proposedZSM_10sats},~\ref{fig:second_proposedZSM_8sats} and~\ref{fig:second_proposedZSM_20elev}, respectively.
From among ones illustrated earlier in Figure~\ref{fig:second_skyplot}, the satellites with PRNs $1,7,14$ and $15$ were eliminated for the $10$ random satellites case, while the satellites with PRNs $5,7,8,9,17$ and $28$ were omitted for the $8$ random satellite case. 
Also, for the case with elevation $>20^{\circ}$, only the GPS satellites with PRNs $7$, $13$, $14$, $17$, $28$ and $30$ were considered. 
Intuitively, the reduced number of GPS satellites mimic the cases when a non-ideal LOS/NLOS classifier in a simulation setting, wherein the users can execute the proposed ZSM by leveraging only the GPS satellites exhibiting a high probability of being either LOS or NLOS while omitting the other less certain ones. 
The detailed steps of convergence for our proposed ZSM algorithm for all available $14$ satellites can be viewed here: \url{https://youtu.be/anIh4hd3ikw}.
Similarly, the illustrations of conventional SM with grid sizes of $3~$m, $5~$m, $10~$m and $30~$m are shown in Figure~\ref{fig:second_conventionalSM_3m}, \ref{fig:second_conventionalSM_5m}, \ref{fig:second_conventionalSM_10m}, and \ref{fig:second_conventionalSM_30m}, respectively.

We validate that the proposed ZSM using all available $14$ satellites successfully estimates two disjoint sets that show resemblance with the most-likely position candidates estimated using conventional SM of different grid sizes.
For the conventional SM technique with grid size of $30~$m, we observe one most-likely candidate~(highest visibility score) that showcases a point-valued accuracy of $12~$m and $60~$m in cross-street and along-street directions, respectively.
By implementing the conventional SM technique for a grid size of $10~$m, we identify two most-likely candidates, both of which exhibit point-valued estimation error in cross-street direction of $2~$m. 
Similarly, conventional SM with grid sizes of $5~$m and $3~$m identify $1-3$ grid points with high visibility score of $14$ in the near-vicinity of each of the two disjoint sets estimated via proposed ZSM. 
While increasing the grid resolution reduces the estimated bounds of conventional SM, we observe that these bound values are more than $100~$m across all grid sizes, in both along-street and cross-street directions.
These large uncertainty bounds from conventional SM can be majorly attributed to the use of Gaussian distribution for approximating the non-linear, multi-modal shadow matching distribution.

\begin{table}[H]
\centering
\caption{Performance comparison between ZSM and conventional SM based on a 3-D building map of San Francisco seen in Figure~\ref{fig:second_3Dsetup} and fourteen GPS satellites seen in Figure~\ref{fig:second_skyplot}.
For proposed ZSM, we perform comparison analysis across different GPS satellite subsets, while for conventional SM we compare the performance for different grid resolutions. }
\begin{tabular}{|cc|c|c|cc|}
\hline
\multicolumn{2}{|c|}{\multirow{3}{*}{\textbf{Algorithm}}} & \multirow{3}{*}{\textbf{\begin{tabular}[c]{@{}c@{}}Centroid error w.r.t true \\ location (m) {[}Cross-street, \\ Along-street{]}\end{tabular}}} & \multirow{3}{*}{\textbf{\begin{tabular}[c]{@{}c@{}}Bounds (m)\\ {[}Cross-street, \\ Along-street{]}\end{tabular}}} & \multicolumn{2}{c|}{\multirow{2}{*}{\textbf{\begin{tabular}[c]{@{}c@{}}Average computation load\\ across 100 runs (s)\end{tabular}}}} \\
\multicolumn{2}{|c|}{} &  &  & \multicolumn{2}{c|}{} \\ \cline{5-6} 
\multicolumn{2}{|c|}{} &  &  & \multicolumn{1}{c|}{\textbf{Offline}} & \textbf{Online} \\ \hline
\multicolumn{1}{|c|}{\multirow{4}{*}{\textbf{\begin{tabular}[c]{@{}c@{}}Proposed \\ ZSM\end{tabular}}}} & \textbf{14 satellites} &  \begin{tabular}[c]{@{}c@{}}{[}\textbf{1.2, 8.1}{]}\\ {[}4.6, 58.8{]}\end{tabular} & \begin{tabular}[c]{@{}c@{}}{[}\textbf{5.8, 16.2}{]}\\ {[}9.8, 24.5{]}\end{tabular}  & \multicolumn{1}{c|}{\textbf{162.5}} & 4.4 \\ \cline{2-6} 
\multicolumn{1}{|c|}{} & \textbf{\begin{tabular}[c]{@{}c@{}}10 random \\ satellites\end{tabular}} &  \begin{tabular}[c]{@{}c@{}}{[}0.9, 6.8{]}\\ {[}4.8, 53.8{]}\end{tabular} & \begin{tabular}[c]{@{}c@{}}{[}6.1, 16.4{]}\\ {[}9.7, 13.1{]}\end{tabular}  & \multicolumn{1}{c|}{\textbf{162.5}} &  4.2 \\ \cline{2-6} 
\multicolumn{1}{|c|}{} & \textbf{\begin{tabular}[c]{@{}c@{}}8 random \\ satellites\end{tabular}} &  \begin{tabular}[c]{@{}c@{}}{[}1.4, 13.1{]}\\ {[}3.5, 54.3{]}\end{tabular} & \begin{tabular}[c]{@{}c@{}}{[}18.3, 26.8{]}\\ {[}7.6, 24.8{]}\end{tabular}  & \multicolumn{1}{c|}{\textbf{162.5}} &  3.0 \\ \cline{2-6} 
\multicolumn{1}{|c|}{} & \textbf{\begin{tabular}[c]{@{}c@{}}Satellites with \\ elevation $>20^\circ$\end{tabular}} &  \begin{tabular}[c]{@{}c@{}}{[}1.7, 10.2{]}\\ {[}1.9, 7.0{]}\\ {[}4.9, 48.5{]}\end{tabular} & \begin{tabular}[c]{@{}c@{}}{[}0.5, 3.3{]}\\ {[}8.7, 20.2{]}\\ {[}16.8, 29.9{]}\end{tabular}  & \multicolumn{1}{c|}{\textbf{162.5}} &  2.4 \\ \hline
\multicolumn{1}{|c|}{\multirow{4}{*}{\textbf{\begin{tabular}[c]{@{}c@{}}Conventional \\ SM\\ (all available \\ 14 satellites)\end{tabular}}}} & \textbf{3 m} & \begin{tabular}[c]{@{}c@{}}{[}0.0, 3.0{]}, {[}0.0, 6.0{]}, {[}0.0, 9.0{]}\\ {[}0.0, 12.0{]}, {[}3.0, 12.0{]}, {[}0.0, 48.0{]}\\ {[}3.0, 48.0{]}, {[}0.0, 51.0{]}, {[}3.0, 51.0{]}\\ {[}0.0, 54.0{]}, {[}3.0, 54.0{]}, {[}3.0, 57.0{]}\end{tabular}  & [173.2, 214.5] & \multicolumn{1}{c|}{91204.0} & 2.8 \\ \cline{2-6} 
\multicolumn{1}{|c|}{} & \textbf{5 m} &  \begin{tabular}[c]{@{}c@{}}{[}2.0,10.0{]} \\ {[}2.0, 50.0{]}, {[}3.0, 50.0{]}, {[}3.0, 55.0{]}\end{tabular} & [173.1, 215.6]  & \multicolumn{1}{c|}{34389.9} & 1.2 \\ \cline{2-6} 
\multicolumn{1}{|c|}{} & \textbf{10 m} &  \begin{tabular}[c]{@{}c@{}}{[}2.0, 10.0{]}\\ {[}2.0, 50.0{]}\end{tabular} & {[}192.1, 221.6{]}  & \multicolumn{1}{c|}{6874.1} & 0.3 \\ \cline{2-6} 
\multicolumn{1}{|c|}{} & \textbf{30 m} &  {[}12.0, 60.0{]} & {[}224.5, 261.8{]}  & \multicolumn{1}{c|}{1433.4} & \textbf{0.1} \\ \hline
\end{tabular}
\label{table:second_stats}
\end{table}
\vspace{-5mm}
In contrast, across all satellite subsets, our proposed ZSM successfully estimates the set containing the true receiver location as one of the outputs, without requiring discretizing. 
Particularly, when all the $14$ available satellites are utilized, ZSM demonstrates a higher point-valued accuracy of $1.2~$m in cross-street and $8.1~$m in along-street direction, while achieving a small bound of only $5.8~$m in cross-street and $16.2~$m in along-street directions. 
We observe that as the number of GPS satellites utilized in our proposed ZSM increases, the centroid error and the bounds of the mode that contains the true receiver location marginally decreases. 
Similarly, as the number of utilized GPS satellites increase, the offline computation load remains the same, but the online computation load marginally increases.
An interesting thing to note here is that while ZSM estimates multiple disjoint components~(modes), the ones that do not contain the true receiver location can be ruled out by fusing proposed ZSM with other information sources, say Inertial Measurement Unit~(IMU) or GPS pseudoranges.
In particular, these additional sources provide temporal information about the vehicle motion that can be used to perform consistency checks, i.e., compute the predicted/expected modes from IMU/GPS pseudoranges, and compare it with the estimated one from zonotope shadow matching to eliminate the incorrect modes over time. 
For instance, IMU data can be utilized to perform consistency checks over multiple time instants in relative domain itself while considering the IMU biases and drifts to be constant over relatively shorter time window. 
Similarly, the GPS pseudoranges from satellites identified as LOS with high certainty can provide useful information, particularly in resolving the along-street component, thereby narrowing down the correct mode. 
Also, note that parallelization can further reduce the computation load of both the proposed ZSM and the conventional SM.
\subsubsection{Sensitivity Analysis of ZSM} \label{subsec:sensitivityanalysis}
We perform sensitivity analysis of ZSM to evaluate the increase in offline and online computation load as the level-of-detail for each building is increased or alternatively the number of buildings is increased. 
As explained earlier in Section~\ref{subsubsec:3-D_map_vertex_rep}, the standard 3-D maps can be represented as a union of triangles, wherein each triangle is converted into a constrained zonotope. 
Therefore, in theory, the computational load incurred by increasing the number of buildings (with sparser level-of-detail for each building) is on the same order of magnitude as increasing the level-of-detail for each building, given a lower number of buildings. 
This is justified by comparing Eqs.~\eqref{eq:assum_buildings_are_conzonos} and~\eqref{eq:building_are_triangles} as follows: a) given $n\trng$ triangles and $n_i=n\trng$, the total number of constrained zonotopes representing the 3-D buildings are $n\trng$~(each building is sparser in the level-of-detail); and b) given another $n\trng$ triangles with $n_i<n\trng$, the total number of constrained zonotopes representing the 3-D buildings are $n_i$~(each building encompasses more level-of-detail).
Given that the open-source 3-D building map~\citep{gyorgy2018sfmap} is not dense enough to conduct sensitivity analysis on level-of-detail explicitly, we analyze the performance of the proposed ZSM as the number of 3-D buildings are varied. 
We also quantify our ZSM's performance by analyzing the following: a) centroid error and bounds associated with the disjoint component~(ambiguous mode) containing the true receiver location; and b) number of ambiguous modes and their variation as the number of 3-D buildings increases.

\begin{table}[H]
\centering
\caption{Sensitivity analysis of ZSM by varying the number of buildings as $8$, $14$ and $20$ indicates that our proposed ZSM not only maintains a high point-valued estimation accuracy but also a small position bound. }
\begin{tabular}{|c|c|c|c|}
\hline
\multirow{2}{*}{\textbf{\begin{tabular}[c]{@{}c@{}}No. of \\ buildings\end{tabular}}} & \multirow{2}{*}{\textbf{\begin{tabular}[c]{@{}c@{}}Centroid Error @ Bound (m)\\ {[}Cross-street, Along-street{]}\end{tabular}}} & \multicolumn{2}{c|}{\textbf{\begin{tabular}[c]{@{}c@{}}Avg. computation load\\ for 100 runs (s)\end{tabular}}} \\ \cline{3-4} 
 &  & \textbf{Offline} & \textbf{Online} \\ \hline
\textbf{8} & \begin{tabular}[c]{@{}c@{}}a) {[}\textbf{1.2, 8.1}{]} @ {[}\textbf{5.8, 16.2}{]}\\ b) {[}4.6, 58.8{]} @ {[}9.8, 24.5{]}\end{tabular} & \textbf{162.5} & \textbf{4.6} \\ \hline
\textbf{14} & \begin{tabular}[c]{@{}c@{}}a) {[}\textbf{1.2, 8.1}{]} @ {[}\textbf{5.8, 16.2}{]}\\ b) {[}4.6, 58.8{]} @ {[}9.8, 24.5{]}\\ c) {[}102.4, 63.9{]} @ {[}4.9, 25.2{]}\end{tabular} & 165.4 & 9.4 \\ \hline
\textbf{20} & \begin{tabular}[c]{@{}c@{}}a) {[}\textbf{1.2, 8.1}{]} @ {[}\textbf{5.8, 16.2}{]}\\b) {[}4.6, 58.8{]} @ {[}9.8, 24.5{]}\\ c) {[}102.4, 63.9{]} @ {[}4.9, 25.2{]}\\ d) {[}1.7, 10.1{]} @ {[}0.4, 3.3{]}\\ e) {[}87.0, 155.9{]} @ {[}0.3, 0.5{]}\end{tabular} & 169.8 & 12.2 \\ \hline
\end{tabular}
\label{table:sensitivity}
\end{table}

\begin{figure}[ht]
    \centering
    \includegraphics[width=0.42\textwidth]{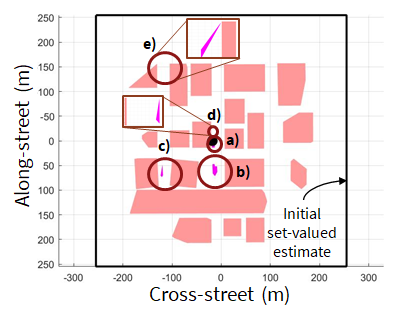}
    \caption{Performance of our proposed ZSM that considers $20$ buildings from the 3-D map of San Francisco. The initial AOI is shown by the black box, while the final set-valued receiver position estimate is represented in magenta. We observe our final position estimate comprises five disjoint components, of which one component contains the true receiver location. The component containing the true receiver location exhibits not only a high accuracy of $1.2~$m and $8.1~$m in cross-street and along-street directions, respectively, but also a small position bound of $5.8~$m and $16.2~$m in cross-street and along-street directions, respectively.  }
    \label{fig:sensitivity}
\end{figure}

As shown in Table~\ref{table:sensitivity}, we analyze our ZSM's performance by increasing the number of buildings from $8$~(discussed in Section~\ref{subsec:second_simulation}) to $14$ and $20$.
For all three cases, we consider the same initial AOI, which is indicated by the black box shown in Figure~\ref{fig:sensitivity}.
Figure~\ref{fig:sensitivity} illustrates the final set-valued position estimate obtained for the case with $20$ buildings.
We validate that our proposed ZSM successfully detects all the disjoint components~(ambiguous modes) and their exact bounds, which can vary based on the number, density and placement of 3-D buildings considered.
From Table~\ref{table:sensitivity}, we observe that the number of disjoint components in the final set-valued position estimate increases as the number of buildings increases, wherein the number of disjoint components is $2$ when $8$ buildings are considered, $3$ disjoint components for $14$ buildings and $5$ disjoint components when $20$ buildings are considered. 
As explained earlier, disjoint components not containing the true receiver location can be omitted by fusing proposed ZSM with other sources, such as IMU and GPS pseudoranges.
We demonstrate that our proposed ZSM successfully maintains a high estimation accuracy and a small position bound in both cross-street and along-street directions.
Another interesting point to note is that the $8$ buildings and $14$ satellites configuration already provides sufficient geometric diversity, and thus, increasing the number of buildings to $14$ and $20$ does not further reduce the centroid error and bound of the ambiguous mode containing the true receiver location. 

Based on the results, we observe that the offline computation load (even with no parallelization as of now) demonstrates easily-scalability with number of 3-D buildings, i.e., no significant increase in computation time. 
While online computation load increases with the number of 3-D buildings, note that there is an extensive scope for parallelization in the extraction of GPS shadows that can drastically reduce the compute time. 
For instance, one can parallelize the processing of analyzing each pair of constrained zonotope~(associated with 3-D buildings) and satellite to extract the GPS shadow, which is currently executed in a sequential manner as explained in Algorithm~\ref{alg:zono_shadow_matching}.
However, note that parallelization incurs an increase in the computational resources required. 

\subsubsection{Performance Analysis of ZSM: Different Location} \label{subsec:third_simulation}

\begin{figure}[ht]
	\setlength{\belowcaptionskip}{-4pt}
	\centering
	\begin{subfigure}[b]{0.45\textwidth}
	    \centering
	    \includegraphics[width=0.98\textwidth]{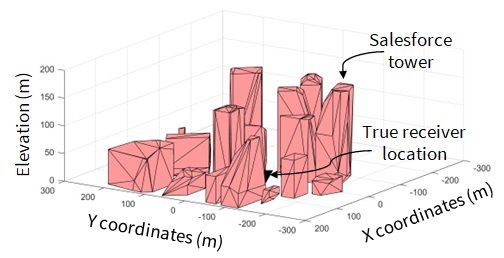}
	\caption{A relevant section from 3-D San Francisco map}
	\label{fig:third_buildingmap}
	\end{subfigure}
	\vspace{5mm}
	\begin{subfigure}[b]{0.4\textwidth}
	    \centering
	    \includegraphics[width=0.98\textwidth]{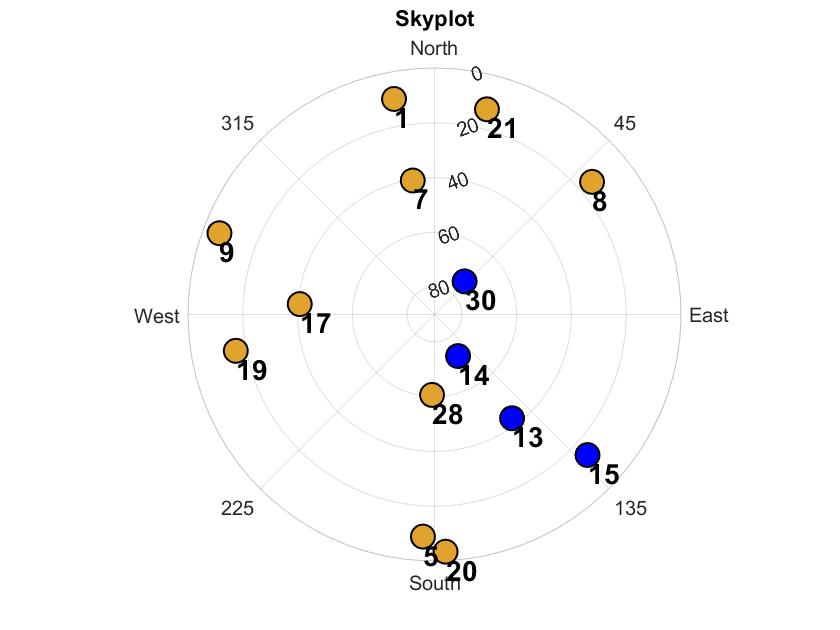}
	\caption{GPS satellite skyplot}
	\label{fig:third_skyplot}
	\end{subfigure}
	\begin{subfigure}[b]{0.4\textwidth}
	    \centering
		\includegraphics[width=0.98\textwidth]{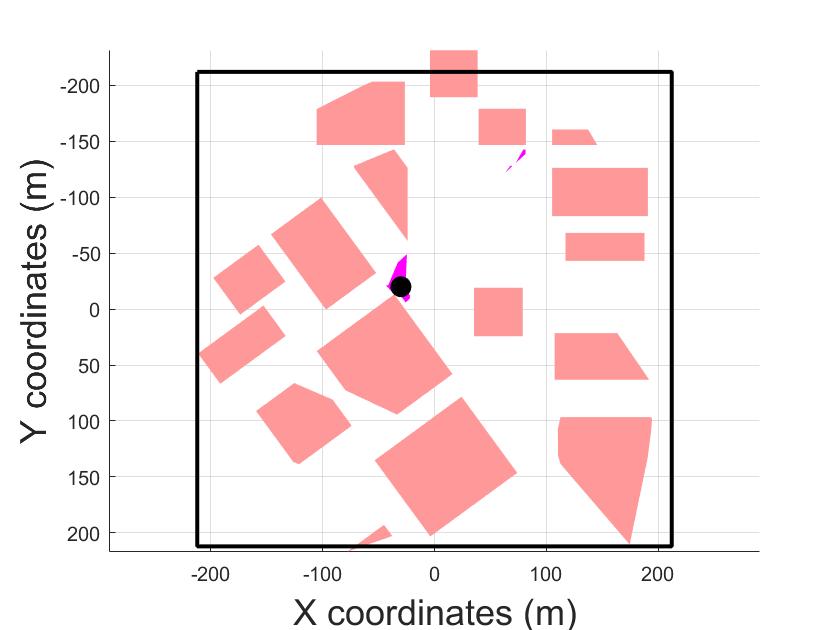}
	\caption{ZSM with all $14$ available satellites}
	\label{fig:third_finalpos}
	\end{subfigure}
	\begin{subfigure}[b]{0.4\textwidth}
	    \centering
		\includegraphics[width=0.98\textwidth]{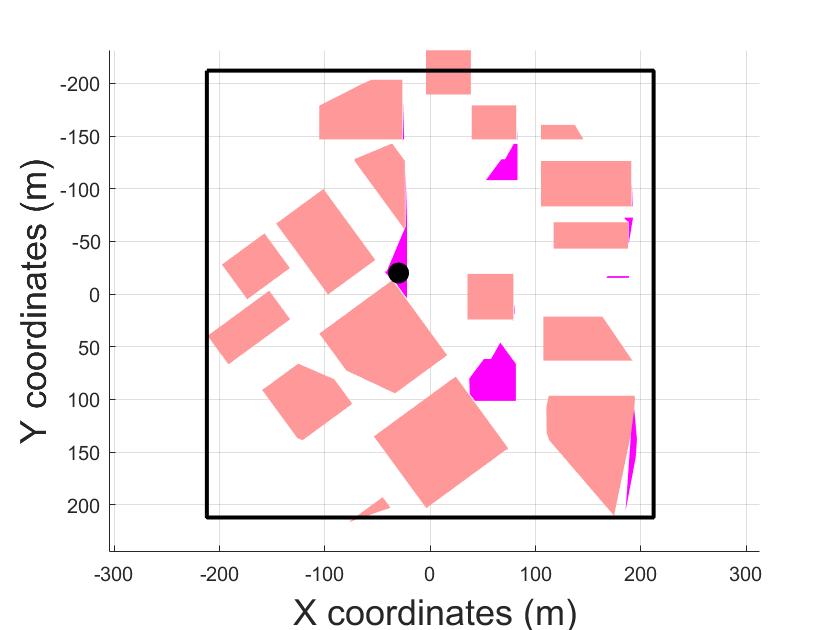}
	\caption{ZSM with satellites above elevation of $20^{\circ}$}
	\label{fig:third_finalposelev20}
	\end{subfigure}
	\caption{Our third simulation experiment in a different location using 3-D map of San Francisco. Subfigure (a) shows a relevant section of 3-D map near the Salesforce tower wherein the distinction between cross-street and along-street is not well-defined.
	Subfigure (b) shows the skyplot with respect to true receiver location with fourteen visible GPS satellites. The GPS satellites are represented by circles with LOS satellites in blue and NLOS (simulated effects) in dark yellow.
	Subfigure (c) shows the final set-valued estimate of receiver position estimated by our ZSM when all $14$ satellites are considered, while Subfigure (d) shows this set-valued estimate using satellites with elevation~$>20^{\circ}$. 
	We demonstrate that in both cases, our ZSM successfully detects the mode containing the true receiver location, while exhibiting a higher centroid error and bounds for the reduced $6$ satellite case as compared to the full $14$ satellite case.}
\end{figure}

We evaluate our ZSM's performance in a different section of the 3-D building map of San Francisco, i.e., near an intersection, where the distinction between the cross-street and along-street directions is not obvious. 
Given this, we label the directions as X and Y in the local map coordinates as shown in Figure~\ref{fig:third_buildingmap}, which comprises $17$ urban buildings in total.
The true receiver position is initialized at (-20, -30) m in local map coordinates and the initial AOI~(black box) is chosen to lie within a size of $212.1$~m $\times 212.1~$m  while excluding the regions that lie within the building footprints.
We consider fourteen GPS satellites, which are same as those considered in earlier subsections and whose skyplot is shown in Figure~\ref{fig:third_skyplot} with LOS satellites~(a total of $4$) indicated by blue circle markers and NLOS by dark yellow.
A visualization of our ZSM with all available $14$ satellites and satellites with elevation~$>20^{\circ}$ are shown in Figures~\ref{fig:third_finalpos} and~\ref{fig:third_finalposelev20}, respectively.
Among all the GPS satellites shown in Figure~\ref{fig:third_skyplot}, for the case with elevation $>20^{\circ}$, only the relevant $6$ satellites with PRNs $7$, $13$, $14$, $17$, $28$ and $30$ are considered. 

We demonstrate that our proposed ZSM estimates all the disjoint components~(ambiguous modes), wherein for the case when all the GPS satellites are considered, $3$ disjoint ones are identified with an online computation load of $10.9~$s~(no parallelization used).
Similarly, for the case comprising of only the GPS satellites with elevation $>20^{\circ}$, $13$ disjoint ones are estimated using our proposed ZSM with an online computation load of $5.5~$s. 
Also observe that in both the cases, our proposed ZSM successfully detects the mode that contains the true receiver location with a low centroid error of $[0.9, 6.0]$~m and a bound of $[21.4, 55.1]$~m in local map coordinates for the $14$ GPS satellite case. 
On the other hand, a higher centroid error of $[1.6, 10.8]$~m and bound of $[21.4, 122.5]$~m in local map coordinates is observed for the $6$ GPS satellite case. 
Unlike in Section~\ref{subsec:second_simulation} where increasing the number of satellites does not significantly affect the characteristics of the mode containing the true receiver location, in this case, we observe an improvement in our ZSM's performance.
As explained before, this can be attributed to the further addition of useful geometric constraints in Figure~\ref{fig:third_finalpos} as compared to Figure~\ref{fig:third_finalposelev20} using the GPS shadows extracted from the remaining $8$ satellites.

\section{Conclusions}\label{sec:conclusions_and_future_work}

We presented Zonotope Shadow Matching (ZSM), a novel approach to set-valued position estimation for 3-D map-aided GNSS.
The method is achieved by leveraging constrained zonotopes, a recent advance in polytope representation.
We computed the 3-D buildings and 2-D GNSS shadows using constrained zonotopes, and then refined the coarse set-valued receiver position estimate based on if receiver lies inside or outside the GNSS shadow~(which is judged by $\cno$ values).
Using simulated experiments on a simple 3-D building map and a dense 3-D map of San Francisco, we validated that the proposed ZSM achieves a high point-valued estimation accuracy with high certainty (small position bounds).
Importantly, while achieving a comparable point-valued accuracy as that of the conventional Shadow Matching (SM) technique, ZSM computes a set-valued state estimate that is independent of the point-valued discretization. 
We also demonstrated the easy scalability of ZSM with the number of buildings considered.
This indicates that ZSM is a promising method for providing robustness guarantees for safety-critical GNSS applications, since such set-valued estimates can be used as measurements for set-valued robust estimators.

\section*{acknowledgements}

We want to thank the National Science Foundation~(NSF) and the Air Force Research Laboratory~(AFRL) for funding this work. 
We also want to thank Adyasha Mohanty and Asta Wu for reviewing our paper. 

\printbibliography[heading=bibintoc, title={References}]

@book{Hofmann_Wellenhof_1992,
	doi = {10.1007/978-3-7091-5126-6},
	url = {https://doi.org/10.1007\%2F978-3-7091-5126-6},
	year = 1992,
	publisher = {Springer Vienna},
	author = {Bernhard Hofmann-Wellenhof and Herbert Lichtenegger and James Collins},
	title = {Global Positioning System}
}

@article{Zhu_2018,
	doi = {10.1109/tits.2017.2766768},
	url = {https://doi.org/10.1109\%2Ftits.2017.2766768},
	year = 2018,
	month = sep,
	publisher = {Institute of Electrical and Electronics Engineers ({IEEE})},
	volume = {19},
	number = {9},
	pages = {2762--2778},
	author = {Ni Zhu and Juliette Marais and David Betaille and Marion Berbineau},
	title = {{GNSS} Position Integrity in Urban Environments: A Review of Literature},
	journal = {{IEEE} Transactions on Intelligent Transportation Systems}
}

@inproceedings{Suzuki_2016,
	doi = {10.33012/2016.14857},
	url = {https://doi.org/10.33012\%2F2016.14857},
	year = 2016,
	month = nov,
	publisher = {Institute of Navigation},
	author = {Taro Suzuki},
	title = {Integration of {GNSS} Positioning and 3D Map using Particle Filter},
	booktitle = {Proceedings of the 29th International Technical Meeting of The Satellite Division of the Institute of Navigation ({ION} {GNSS}+ 2016)}
}

@inproceedings{Miura_2013,
	doi = {10.1109/itsc.2013.6728447},
	url = {https://doi.org/10.1109\%2Fitsc.2013.6728447},
	year = 2013,
	month = oct,
	publisher = {{IEEE}},
	author = {Shunsuke Miura and Shoma Hisaka and Shunsuke Kamijo},
	title = {{GPS} multipath detection and rectification using 3D maps},
	booktitle = {16th International {IEEE} Conference on Intelligent Transportation Systems ({ITSC} 2013)}
}

@article{Miura_2015,
	doi = {10.1109/tits.2015.2432122},
	url = {https://doi.org/10.1109\%2Ftits.2015.2432122},
	year = 2015,
	month = dec,
	publisher = {Institute of Electrical and Electronics Engineers ({IEEE})},
	volume = {16},
	number = {6},
	pages = {3104--3115},
	author = {Shunsuke Miura and Li-Ta Hsu and Feiyu Chen and Shunsuke Kamijo},
	title = {{GPS} Error Correction With Pseudorange Evaluation Using Three-Dimensional Maps},
	journal = {{IEEE} Transactions on Intelligent Transportation Systems}
}

@misc{iland_uber_rethinking_gps,
    title = {{Rethinking GPS: Engineering Next-Gen Location at Uber}},
    author = {Iland, Danny and Irish, Andrew and Madhow, Upamanyu and Sandler, Brian},
    year = {2018}, 
    note = {Available: \url{https://eng.uber.com/rethinking-gps/} (Accessed 04/2022)}
}

@misc{van_diggelen_android_improving_urban_gps,
    title = {{Improving urban GPS accuracy for your app}},
    author = {Van Diggelen, Frank and Wang, Jennifer},
    year = {2020},
    journal={Android Developers Blog},
    note={Available: \url{https://android-developers.googleblog.com/2020/12/improving-urban-gps-accuracy-for-your.html} (Accessed 04/22)}
}

@article{Wang_2013,
	doi = {10.1002/navi.38},
	url = {https://doi.org/10.1002\%2Fnavi.38},
	year = 2013,
	month = aug,
	publisher = {Wiley},
	volume = {60},
	number = {3},
	pages = {195--207},
	author = {Lei Wang and Paul D. Groves and Marek K. Ziebart},
	title = {{GNSS} Shadow Matching: Improving Urban Positioning Accuracy Using a 3D City Model with Optimized Visibility Scoring Scheme},
	journal = {Navigation}
}

@article{Groves_2011,
	doi = {10.1017/s0373463311000087},
	url = {https://doi.org/10.1017\%2Fs0373463311000087},
	year = 2011,
	month = jun,
	publisher = {Cambridge University Press ({CUP})},
	volume = {64},
	number = {3},
	pages = {417--430},
	author = {Paul D. Groves},
	title = {Shadow Matching: A New {GNSS} Positioning Technique for Urban Canyons},
	journal = {Journal of Navigation}
}

@inproceedings{wang2013urban,
  title={\href{https://discovery.ucl.ac.uk/id/eprint/1394970/}{Urban positioning on a smartphone: Real-time shadow matching using GNSS and 3D city models}},
  author={Wang, Lei and Groves, Paul D and Ziebart, Marek K},
  year={2013},
  organization={The Institute of Navigation},
  booktitle={Proceedings of the 26th International Technical Meeting of The Satellite Division of the Institute of Navigation (ION GNSS+ 2013)}
}

@inproceedings{groves2015shadow_matching_challenges,
  title={\href{https://discovery.ucl.ac.uk/id/eprint/1472568/}{GNSS shadow matching: The challenges ahead}},
  author={Groves, Paul D and Wang, Lei and Adjrad, Mounir and Ellul, Claire},
  year={2015},
  organization={The Institute of Navigation},
  booktitle={Proceedings of the 28th International Technical Meeting of The Satellite Division of the Institute of Navigation (ION GNSS+ 2015)}
}

@article{Wang_2014,
	doi = {10.1017/s0373463314000836},
	url = {https://doi.org/10.1017\%2Fs0373463314000836},
	year = 2014,
	month = dec,
	publisher = {Cambridge University Press ({CUP})},
	volume = {68},
	number = {3},
	pages = {411--433},
	author = {Lei Wang and Paul D Groves and Marek K Ziebart},
	title = {Smartphone Shadow Matching for Better Cross-street {GNSS} Positioning in Urban Environments},
	journal = {Journal of Navigation}
}

@article{Shi_2015,
	doi = {10.1109/tac.2014.2370472},
	url = {https://doi.org/10.1109\%2Ftac.2014.2370472},
	year = 2015,
	month = may,
	publisher = {Institute of Electrical and Electronics Engineers ({IEEE})},
	volume = {60},
	number = {5},
	pages = {1275--1290},
	author = {Dawei Shi and Tongwen Chen and Ling Shi},
	title = {On Set-Valued Kalman Filtering and Its Application to Event-Based State Estimation},
	journal = {{IEEE} Transactions on Automatic Control}
}

@article{Shiryaev_2015,
	doi = {10.1016/j.proeng.2015.12.045},
	url = {https://doi.org/10.1016\%2Fj.proeng.2015.12.045},
	year = 2015,
	publisher = {Elsevier {BV}},
	volume = {129},
	pages = {252--258},
	author = {V.I. Shiryaev and E.O. Podivilova},
	title = {Set-valued Estimation of Linear Dynamical System State When Disturbance is Decomposed as a System of Functions},
	journal = {Procedia Engineering}
}

@inproceedings{Combettes_1991,
	doi = {10.1109/icassp.1991.151014},
	url = {https://doi.org/10.1109\%2Ficassp.1991.151014},
	year = 1991,
	publisher = {{IEEE}},
	author = {P.L. Combettes and M.R. Civanlar},
	title = {The foundations of set theoretic estimation},
	booktitle = {[Proceedings] {ICASSP} 91: 1991 International Conference on Acoustics, Speech, and Signal Processing}
}

@article{Scott_2016,
	doi = {10.1016/j.automatica.2016.02.036},
	url = {https://doi.org/10.1016\%2Fj.automatica.2016.02.036},
	year = 2016,
	month = jul,
	publisher = {Elsevier {BV}},
	volume = {69},
	pages = {126--136},
	author = {Joseph K. Scott and Davide M. Raimondo and Giuseppe Roberto Marseglia and Richard D. Braatz},
	title = {Constrained zonotopes: A new tool for set-based estimation and fault detection},
	journal = {Automatica}
}

@inproceedings{Shetty_2020,
	doi = {10.33012/2020.17518},
	url = {https://doi.org/10.33012\%2F2020.17518},
	year = 2020,
	month = oct,
	publisher = {Institute of Navigation},
	author = {Akshay Shetty and Grace Xingxin Gao},
	title = {Trajectory Planning Under Stochastic and Bounded Sensing Uncertainties Using Reachability Analysis},
	booktitle = {Proceedings of the 33rd International Technical Meeting of the Satellite Division of The Institute of Navigation ({ION} {GNSS}+ 2020)}
}

@inproceedings{Bhamidipati_2020,
	doi = {10.33012/2020.17546},
	url = {https://doi.org/10.33012\%2F2020.17546},
	year = 2020,
	month = oct,
	publisher = {Institute of Navigation},
	author = {Sriramya Bhamidipati and Grace Xingxin Gao},
	title = {Integrity-Driven Landmark Attention for {GPS}-Vision Navigation via Stochastic Reachability},
	booktitle = {Proceedings of the 33rd International Technical Meeting of the Satellite Division of The Institute of Navigation ({ION} {GNSS}+ 2020)}
}

@article{neamati2022mosaic,
  title={Mosaic Zonotope Shadow Matching for Risk-Aware Autonomous Localization in Harsh Urban Environments},
  author={Neamati, Daniel and Bhamidipati, Sriramya and Gao, Grace},
  journal={Journal of Artificial Intelligence. [Submitted. Available: \url{https://stanford.box.com/shared/static/r0abx08okk5do2gbuhclhr1fuc4ayvij.pdf}]},
  year={2022},
  note={}
}

@inproceedings{Neamati_2022,
	year = 2022,
	month = sep,
	publisher = {Institute of Navigation},
	author = {Daniel Neamati and Sriramya Bhamidipati and Grace Xingxin Gao},
	title = {Set-Based Ambiguity Reduction in Shadow Matching with Iterative GNSS Pseudoranges},
	booktitle = {Proceedings of the 35th International Technical Meeting of the Satellite Division of The Institute of Navigation ({ION} {GNSS}+ 2022) [Abstract Accepted. Available at: \url{https://www.ion.org/gnss/abstracts.cfm?paperID=11326}]}
}

@inproceedings{Xu2018classifier,
  title = {{{GNSS Shadow Matching}} Based on {{Intelligent LOS}}/{{NLOS Classifier}}},
  booktitle = {The 16th World Congress of the {{International Association}} of {{Institutes}} of {{Navigation}} ({{IAIN}})},
  author = {Xu Haosheng and Zhang Guohao and Xu Bing and Hsu Li-Ta},
  year = {2018},
  month = dec,
  pages = {1-7},
  address = {{Chiba, Japan}}
}

@article{xu2020machine,
  doi = {10.1186/s43020-020-00016-w},
  title={Machine learning based {LOS}/{NLOS} classifier and robust estimator for {GNSS} shadow matching},
  author={Xu Haosheng and Angrisano Antonio and Gaglione Salvatore and Hsu Li-Ta},
  journal={Satellite Navigation},
  volume={1},
  pages={1--12},
  year={2020},
  publisher={Springer}
}

@inproceedings{Kousik_2019,
	doi = {10.1115/dscc2019-9214},
	url = {https://doi.org/10.1115\%2Fdscc2019-9214},
	year = 2019,
	month = oct,
	publisher = {American Society of Mechanical Engineers},
	author = {Shreyas Kousik and Patrick Holmes and Ram Vasudevan},
	title = {Safe, Aggressive Quadrotor Flight via Reachability-Based Trajectory Design},
	booktitle = {Volume 3, Rapid Fire Interactive Presentations: Advances in Control Systems: Advances in Robotics and Mechatronics: Automotive and Transportation Systems: Motion Planning and Trajectory Tracking: Soft Mechatronic Actuators and Sensors: Unmanned Ground and Aerial Vehicles}
}

@article{Althoff_2014,
	doi = {10.1109/tro.2014.2312453},
	url = {https://doi.org/10.1109\%2Ftro.2014.2312453},
	year = 2014,
	month = aug,
	publisher = {Institute of Electrical and Electronics Engineers ({IEEE})},
	volume = {30},
	number = {4},
	pages = {903--918},
	author = {Matthias Althoff and John M. Dolan},
	title = {Online Verification of Automated Road Vehicles Using Reachability Analysis},
	journal = {{IEEE} Transactions on Robotics}
}

@inproceedings{althoff2016cora,
	author = {Althoff, M.},
	title = {An Introduction to {CORA} 2015},
	booktitle = {In Proceedings of the Workshop on Applied Verification for Continuous and Hybrid Systems},
	year = {2015},
	pages = {120–151},
	doi = {10.29007/zbkv},
}

@article{raghuraman2020_cons_zono_ops,
  title={\href{https://arxiv.org/abs/2009.06039}{Set operations and order reductions for constrained zonotopes}},
  author={Raghuraman, Vignesh and Koeln, Justin P},
  journal={arXiv preprint arXiv:2009.06039},
  year={2020}
}

@article{Adjrad_2017,
	doi = {10.1017/s0373463317000509},
	url = {https://doi.org/10.1017\%2Fs0373463317000509},
	year = 2017,
	month = aug,
	publisher = {Cambridge University Press ({CUP})},
	volume = {71},
	number = {1},
	pages = {1--20},
	author = {Mounir Adjrad and Paul D. Groves},
	title = {Intelligent Urban Positioning: Integration of Shadow Matching with 3D-Mapping-Aided {GNSS} Ranging},
	journal = {Journal of Navigation}
}

@misc{vert2lcon,
    author={Matt, Jack},
    title={{Analyze N-dimensional Convex Polyhedra}},
    publisher={{MATLAB Central File Exchange}}, 
    year = {2021},
    note={Available: \url{https://www.mathworks.com/matlabcentral/fileexchange/30892-analyze-n-dimensional-convex-polyhedra} (Accessed 04/2022)}
}

@inproceedings{Herceg_2013,
	doi = {10.23919/ecc.2013.6669862},
	url = {https://doi.org/10.23919\%2Fecc.2013.6669862},
	year = 2013,
	month = jul,
	publisher = {{IEEE}},
	author = {Martin Herceg and Michal Kvasnica and Colin N. Jones and Manfred Morari},
	title = {Multi-Parametric Toolbox 3.0},
	booktitle = {2013 European Control Conference ({ECC})}
}

@article{Martens_2003,
	doi = {10.1002/cem.780},
	url = {https://doi.org/10.1002\%2Fcem.780},
	year = 2003,
	month = mar,
	publisher = {Wiley},
	volume = {17},
	number = {3},
	pages = {153--165},
	author = {Harald Martens and Martin H{\o}y and Barry M. Wise and Rasmus Bro and Per B. Brockhoff},
	title = {Pre-whitening of data by covariance-weighted pre-processing},
	journal = {Journal of Chemometrics}
}

@inproceedings{Fu_2020,
	doi = {10.33012/2020.17628},
	url = {https://doi.org/10.33012\%2F2020.17628},
	year = 2020,
	month = oct,
	publisher = {Institute of Navigation},
	author = {Fu, Guoyu Michael and Mohammed Khider and Frank van Diggelen},
	title = {Android Raw {GNSS} Measurement Datasets for Precise Positioning},
	booktitle = {Proceedings of the 33rd International Technical Meeting of the Satellite Division of The Institute of Navigation ({ION} {GNSS}+ 2020)}
}

@misc{hetet2000signal,
  title={{Signal-to-noise ratio effects on the quality of GPS observations}},
  author={Hetet, Sophie},
  journal={Fredericton, New Brunswick: Department of Geodesy and Geomatics, University of New Brunswick},
  year={2000},
  publisher={Citeseer}, 
  note = {Available: \url{http://gauss.gge.unb.ca/papers.pdf/hetet.report.pdf} (Accessed 04/2022)}
}

@article{Kuusniemi_2004,
	doi = {10.1007/s10291-004-0113-7},
	url = {https://doi.org/10.1007\%2Fs10291-004-0113-7},
	year = 2004,
	month = oct,
	publisher = {Springer Science and Business Media {LLC}},
	volume = {8},
	number = {4},
	pages = {226--237},
	author = {Heidi Kuusniemi and Gerard Lachapelle and Jarmo H. Takala},
	title = {Position and velocity reliability testing in degraded {GPS} signal environments},
	journal = {{GPS} Solutions}
}

@misc{softGNSS,
  author = {Rojas, Cristian Paul Peñaranda},
  title = {Soft {GNSS}},
  year = {2011}, 
  note = {Available: \url{https://github.com/kristianpaul/SoftGNSS} (Accessed 04/2022)}
}

@misc{gpssdrsim,
  author = {Ebinuma, Takuji},
  title = {{GPS}-{SDR}-{SIM}},
  year = {2018}, 
  note = {Available: \url{https://github.com/osqzss/gps-sdr-sim} (Accessed 04/2022)}
}

@inproceedings{Bhamidipati_2019,
	doi = {10.1109/icc.2019.8761208},
	url = {https://doi.org/10.1109\%2Ficc.2019.8761208},
	year = 2019,
	month = may,
	publisher = {{IEEE}},
	author = {Sriramya Bhamidipati and Kyeong Jin Kim and Hongbo Sun and Philip V. Orlik},
	title = {{GPS} Spoofing Detection and Mitigation in {PMUs} using Distributed Multiple Directional Antennas},
	booktitle = {{ICC} 2019 - 2019 {IEEE} International Conference on Communications ({ICC})}
}

@misc{gyorgy2018sfmap,
	author = {György, R},
	title = {{3DWarehouse San Francisco}},
	year = {2018},
	note = {Available: \url{https://3dwarehouse.sketchup.com/model/4ad4796d-8102-4bdd-9bfc-5442dee9facf/San-Francisco?hl=en&login=true} (Accessed 09/2021)}
}

@incollection{K_nig_1990,
	doi = {10.1007/978-1-4684-8971-2_2},
	url = {https://doi.org/10.1007\%2F978-1-4684-8971-2_2},
	year = 1990,
	publisher = {Birkhäuser Boston},
	pages = {45--421},
	author = {D{\'{e}}nes König},
	title = {Theory of Finite and Infinite Graphs},
	booktitle = {Theory of Finite and Infinite Graphs}
}

@inproceedings{Bhamidipati_2021,
	doi = {10.33012/2021.17933},
	url = {https://doi.org/10.33012\%2F2021.17933},
	year = 2021,
	month = oct,
	publisher = {Institute of Navigation},
	author = {Sriramya Bhamidipati and Shreyas Kousik and Grace Gao},
	title = {Set-Valued Shadow Matching using Zonotopes},
	booktitle = {Proceedings of the 34th International Technical Meeting of the Satellite Division of The Institute of Navigation ({ION} {GNSS}+ 2021)}
}

@article{Chen_2017,
	doi = {10.3390/s17010150},
	url = {https://doi.org/10.3390\%2Fs17010150},
	year = 2017,
	month = jan,
	publisher = {{MDPI} {AG}},
	volume = {17},
	number = {12},
	pages = {150},
	author = {Ziyue Chen and Bingbo Gao and Bernard Devereux},
	title = {State-of-the-Art: {DTM} Generation Using Airborne {LIDAR} Data},
	journal = {Sensors}
}

\end{document}